\documentclass{article} % For LaTeX2e
\usepackage{iclr2022_conference,times}

\usepackage{amsmath,amsfonts,bm}

\newcommand{\ptrue}{p}

\newcommand{\ptrain}{p_{\rm{train}}}
\newcommand{\ptest}{p_{\rm{test}}}

\newcommand{\dataset}{\mathcal{D}}
\newcommand{\train}{\mathcal{D_{\mathrm{train}}}}
\newcommand{\valid}{\mathcal{D_{\mathrm{valid}}}}
\newcommand{\test}{\mathcal{D_{\mathrm{test}}}}

\newcommand{\attribute}[1]{y^{#1}}      %
\newcommand{\NumOfAttributes}{K}        %
\newcommand{\attributeSet}[1]{\sA^{#1}} %
\newcommand{\lat}{z}

\def\figref#1{figure~\ref{#1}}

\def\tabref#1{table~\ref{#1}}

\def\multfigref#1#2{figures \ref{#1}-\ref{#2}}

\def\secref#1{section~\ref{#1}}

\def\appref#1{appendix~\ref{#1}}

\def\eqref#1{equation~\ref{#1}}

\def\1{\bm{1}}

\def\vx{{\bm{x}}}

\def\mI{{\bm{I}}}

\DeclareMathAlphabet{\mathsfit}{\encodingdefault}{\sfdefault}{m}{sl}
\SetMathAlphabet{\mathsfit}{bold}{\encodingdefault}{\sfdefault}{bx}{n}

\def\sA{{\mathbb{A}}}

\newcommand{\E}{\mathbb{E}}

\usepackage[colorlinks=true,citecolor=blue]{hyperref}
\usepackage{url}
\usepackage{comment}
\usepackage{soul}
\usepackage{devanagari}
\usepackage{xspace}
\usepackage{graphicx}
\usepackage{caption}
\usepackage{floatrow}
\usepackage{subcaption}
\usepackage{listings}
\usepackage{xcolor}
\usepackage{booktabs}

\usepackage{pifont}% http://ctan.org/pkg/pifont
\newcommand{\cmark}{\ding{51}}%
\newcommand{\xmark}{\ding{55}}%

\definecolor{codegreen}{rgb}{0,0.6,0}
\definecolor{codegray}{rgb}{0.5,0.5,0.5}
\definecolor{codepurple}{rgb}{0.58,0,0.82}
\definecolor{backcolour}{rgb}{0.95,0.95,0.92}

\lstdefinestyle{pythonstyle}{
    backgroundcolor=\color{backcolour},
    commentstyle=\color{codegreen},
    keywordstyle=\color{magenta},
    numberstyle=\tiny\color{codegray},
    stringstyle=\color{codegreen},
    basicstyle=\ttfamily\tiny,
    breakatwhitespace=false,         
    breaklines=true,
    captionpos=b,
    keepspaces=true,
    numbers=left,
    numbersep=5pt,
    showspaces=false,
    showstringspaces=false,
    showtabs=false,
    tabsize=2
}

\definecolor{TartOrange}{HTML}{ff2e35}
\definecolor{Orange}{HTML}{ff7825}
\definecolor{Mango}{HTML}{ffc013}
\definecolor{AppleGreen}{HTML}{7cb81b}
\definecolor{Blue}{HTML}{1173b0}
\definecolor{BdazzledBlue}{HTML}{2e58a5}
\definecolor{Purple}{HTML}{5b3590}
\definecolor{Sunglow}{HTML}{FFCA3A}

\newcommand{\wilds}{\textsc{WILDS}\xspace}
\newcommand{\camelyon}{\textsc{Camelyon17}\xspace}
\newcommand{\dsprites}{\textsc{dSprites}\xspace}
\newcommand{\mpithreed}{\textsc{MPI3D}\xspace}
\newcommand{\smallnorb}{\textsc{SmallNorb}\xspace}
\newcommand{\shapes}{\textsc{Shapes3D}\xspace}
\newcommand{\iwildcam}{\textsc{iWildCam}\xspace}

\newcommand{\stylegan}{\textsc{StyleGAN}\xspace}

\newcommand{\cyclegan}{\textsc{CycleGAN}\xspace}
\newcommand{\stargan}{\textsc{StarGAN}\xspace}

\title{A Fine-Grained Analysis on Distribution Shift}

\author{\bf Olivia Wiles \quad  Sven Gowal \quad  Florian Stimberg \quad  Sylvestre Alvise-Rebuffi \\ 
{\bf Ira Ktena \quad Krishnamurthy (Dj) Dvijotham \quad  Taylan Cemgil}  \\
DeepMind, London, UK \\
\tiny{\texttt{\{oawiles,sgowal,stimberg,sylvestre,iraktena,taylancemgil\}@deepmind.com}} \quad \quad
\tiny{\texttt{dvij@google.com}} \\
}

\iclrfinalcopy % Uncomment for camera-ready version, but NOT for submission.
\begin{document}

\maketitle

\begin{abstract}
Robustness to distribution shifts is critical for deploying machine learning models in the real world.
Despite this necessity, there has been little work in defining the underlying mechanisms that cause these shifts and evaluating the robustness of algorithms across multiple, different distribution shifts.
To this end, we introduce a framework that enables fine-grained analysis of various distribution shifts.
We provide a holistic analysis of current state-of-the-art methods by evaluating 19 distinct methods grouped into five categories across both synthetic and real-world datasets. 
Overall, we train more than 85K models.
Our experimental framework can be easily extended to include new methods, shifts, and datasets.
We find, unlike previous work~\citep{Gulrajani20}, that progress has been made over a standard ERM baseline; in particular, pretraining and augmentations (learned or heuristic) offer large gains in many cases.
However, the best methods are not consistent over different datasets and shifts.
% We will open source our experimental framework, allowing future work to evaluate new methods over multiple shifts to obtain a more complete picture of a method's effectiveness.
\end{abstract}

\section{Introduction}

If machine learning models are to be ubiquitous in critical applications such as driverless cars  \citep{Janai20}, medical imaging \citep{Erickson17}, and science \citep{Jumper21}, it is pivotal to build models that are robust to distribution shifts. 
Otherwise, models may fail surprisingly in ways that derail trust in the system.
For example, \cite{Wilds2020,Perone19,Albadawy18,Heaven20,Castro20} find that a model trained on one set of hospitals may not generalise to the imaging conditions of another; \cite{Alcorn19,Dai18} find that a model for driverless cars may not generalise to new lighting conditions or object poses; and \cite{Buolamwini18} find that a model may perform worse on subsets of the distribution, such as different ethnicities, if the training set has an imbalanced distribution.
Thus, it is important to understand when we expect a model to generalise and when we do not.
This would allow a practitioner to have confidence in the system (e.g.~if a model is demonstrated to be robust to the imaging conditions of different hospitals, then it can be deployed in new hospitals with confidence).

While domain generalization is a well studied area,  \cite{Gulrajani20,Schott21} have cast doubt on the efficacy of existing methods, raising the question: {\em has any progress been made in domain generalization over a standard expectation risk minimization (ERM) algorithm?}
% They demonstrate that supposed state-of-the-art approaches fail to consistently improve upon a standard ERM baseline in a variety of datasets and settings.
Despite these discouraging results, there are many examples that machine learning models \emph{do} generalise across datasets with different distributions. 
For example, CLIP \citep{Radford21}, with well engineered prompts, generalizes to many standard image datasets. 
\cite{Taori20} found that models trained on one image dataset generalise to another, albeit with some drop in performance; in particular, higher performing models generalise better.
However, there is little understanding and experimentation on {\em when} and {\em why} models generalise, especially in realistic settings inspired by real-world applications.
This begs the following question:

\begin{center}
{\em Can we define the important distribution shifts to be robust to  and then systematically evaluate the robustness of different methods?}
\end{center}

To answer the above question, we present a grounded understanding of robustness to distribution shifts.
We draw inspiration from disentanglement literature (see \secref{sec:relatedwork}), which aims to separate images into an independent set of factors of variation (or attributes).
% While this is not pragmatic for real-world data, for which the factors of variation may be numerous and complex, we use the notion of factors of variation (or attributes) to build our robustness framework to describe the problem and evaluation framework to compare different approaches.
In brief, we assume the data is composed of some (possibly extremely large) set of attributes.
We expect models, having seen some distribution of values for an attribute, to be able to learn invariance to that attribute and so to generalise to unseen examples of the attribute and different distributions over that attribute.
Using a simple example to clarify the setup, assume our data has two attributes (shape and color) among others. 
Given data with some distribution over the set of possible colors (e.g.~red and blue) and the task of predicting shape (e.g.~circle or square), we want our model to generalise to unseen colors (e.g.~green) or a different distribution of colors (e.g.~there are very few red circles in the training set, but the samples at evaluation are uniformly sampled from the set of possible colors and shapes).

Using this framework, we evaluate models across three distribution shifts: {\em spurious correlation}, {\em low-data drift}, and {\em unseen data shift} (illustrated in \figref{fig:distribuionshift}) and two additional conditions (label noise and dataset size).
% We evaluate algorithms under the following distribution shifts: (1) attributes are correlated at train but not test time ({\bf spurious correlation}); (2) only a few samples of some attribute values are given at train time but all values are evenly distributed at test time ({\bf low-data drift}); and (3) some attribute values are unseen at train ({\bf unseen data shift}).
We choose these settings as they arise in the real world and harm generalization performance.
Moreover, in our framework, these distribution shifts are the fundamental blocks of building more complex distribution shifts.
We additionally evaluate models when there is varying amounts of label noise (as inspired by noise arising from human raters) and when the total size of the train set varies (to understand how models perform as the number of training examples changes).
The unique ability of our framework to evaluate fine-grained performance of models across different distribution shifts and under different conditions is of critical importance to analyze methods under a variety of real-world settings.
This work makes the following contributions:
\begin{itemize}
    \item We propose a framework to define when and why we expect methods to generalise. We use this framework to define three, real world inspired distribution shifts. We then use this framework to create a systematic evaluation setup across real and synthetic datasets for different distribution shifts.
    Our evaluation framework is easily extendable to new distribution shifts, datasets, or methods to be evaluated.
    \item We evaluate and compare 19 different methods (training more than $85$K models) in these settings. These methods span the following 5 common approaches: architecture choice, data augmentation, domain generalization, adaptive algorithms, and representation learning. This allows for a direct comparison across different areas in machine learning.
    \item We find that simple techniques, such as data augmentation and pretraining are often effective and that domain generalization algorithms do work for certain datasets and distribution shifts. However, there is no easy way to select the best approach a-priori and results are inconsistent over different datasets and attributes, demonstrating there is still much work to be done to improve robustness in real-world settings.
\end{itemize}

\section{Framework to Evaluate Generalization}
\label{sec:robustnessframework}

In this section we introduce our robustness framework for characterizing distribution shifts in a principled manner.
We then relate three common, real world inspired distribution shifts.

\subsection{Latent Factorisation}

We assume a joint distribution $\ptrue$ of inputs $\vx$ and corresponding attributes
$\attribute{1}, \attribute{2}, \dots, \attribute{\NumOfAttributes}$ (denoted as $\attribute{1:\NumOfAttributes}$) with $\attribute{k} \in \attributeSet{k}$ where $\attributeSet{k}$ is a finite set. 
One of these $\NumOfAttributes$ attributes is a label of interest, denoted as $y^l$ (in a mammogram, the label could be cancer/benign and a nuisance attribute $y^i$ with $i \neq l$ could be the identity of the hospital where the mammogram was taken).
Our aim is to build a classifier $f$ that minimizes the risk $R$.
However, in real-world applications, we only have access to a finite set of inputs and attributes of size $n$.
Hence, we minimize the empirical risk $\hat{R}$ instead:
\begin{align*}
   R(f) = \E_{({\vx}, y^l) \sim \ptrue} \left[\mathcal{L}(\attribute{l}, f({\vx})) \right] & &\hat{R}(f; \ptrue) = \frac{1}{n} \sum_{\{(y^l_i, \vx_i) \sim \ptrue\}_{i=1}^n} \mathcal{L}(\attribute{l}_i, f(\vx_i)).
\end{align*}
where $\mathcal{L}$ is a suitable loss function.
Here, all nuisance attributes $\attribute{k}$ with $k\neq l$ are ignored and we work with samples obtained from the marginal $\ptrue(\attribute{l}, \vx)$.
In practice, however, due to selection bias or other confounding factors in data collection, we 
are only able to train and test our models on data collected from two related but distinct distributions: $\ptrain, \ptest$.
For example, $\ptrain$ and $\ptest$ may be concentrated on different subsets of hospitals and this discrepancy  
%We assume that the support of labels in $\ptrain$ is the same as that in $\ptrue$.
may result in a distribution shift; for example, hospitals may use different equipment, leading to different staining on their cell images.
While we train $f$ on data from $\ptrain$ by minimizing $\hat{R}(f; \ptrain)$, we aim to learn a model that generalises well to data from $\ptest$; that is, it should achieve a small $\hat{R}(f; \ptest)$. 

%(and, in practice, other distributions over the same domain as $\ptrue$).
% distributions over images 
% $\ptrain(\vx)$ and $\ptest(\vx)$ where in general $\ptest(\vx) \neq \ptrain(\vx) \neq \ptrue(\vx)$.
% For example, $\ptrain$ consists of data from a subset of hospitals in $\ptrue$ and $\ptest$ from a different subset.
% We also don't have direct access to $\ptrain$ or $\ptest$, but have two datasets $\train $ and $\test$, where each image $\vx \in \train$ is identically and independently drawn $\vx \sim \ptrain$ (similarly 
% for $\vx \in \test$ drawn as $\vx \sim \ptest$).

\begin{figure}
    \begin{subfigure}[b]{0.24\linewidth} 
    \footnotesize
    \includegraphics[width=\linewidth]{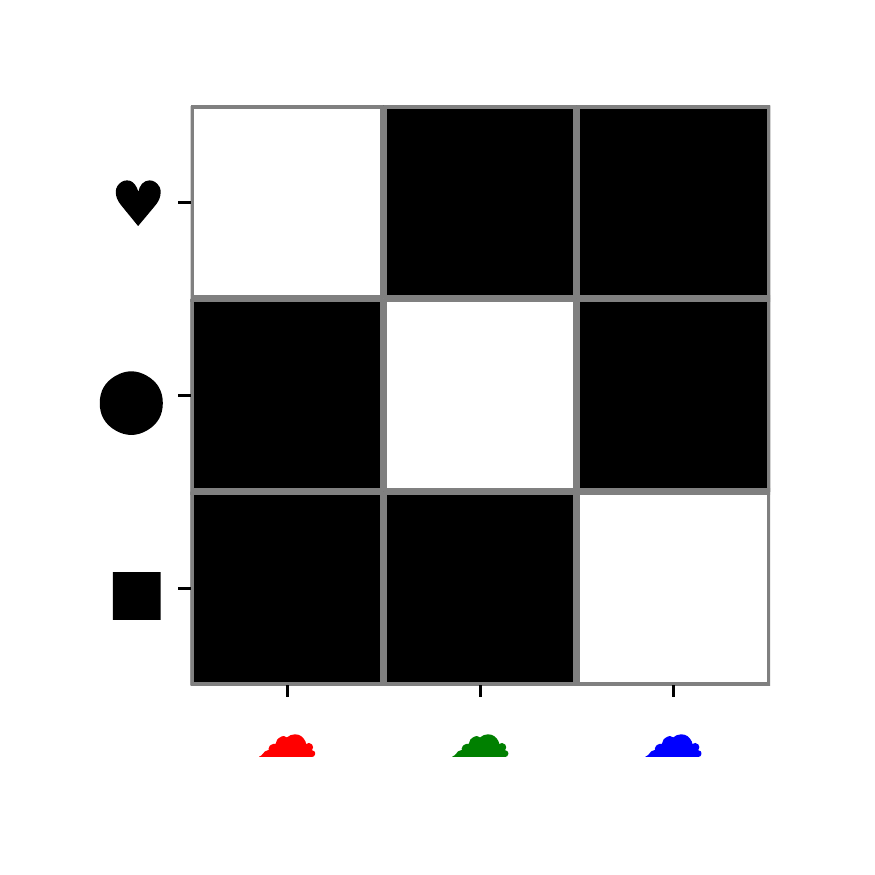}
    \caption{{\bf $\ptrain$: SC.}}
    \label{fig:shift:SC}
    \end{subfigure}
    \begin{subfigure}[b]{0.24\linewidth}
    \includegraphics[width=\linewidth]{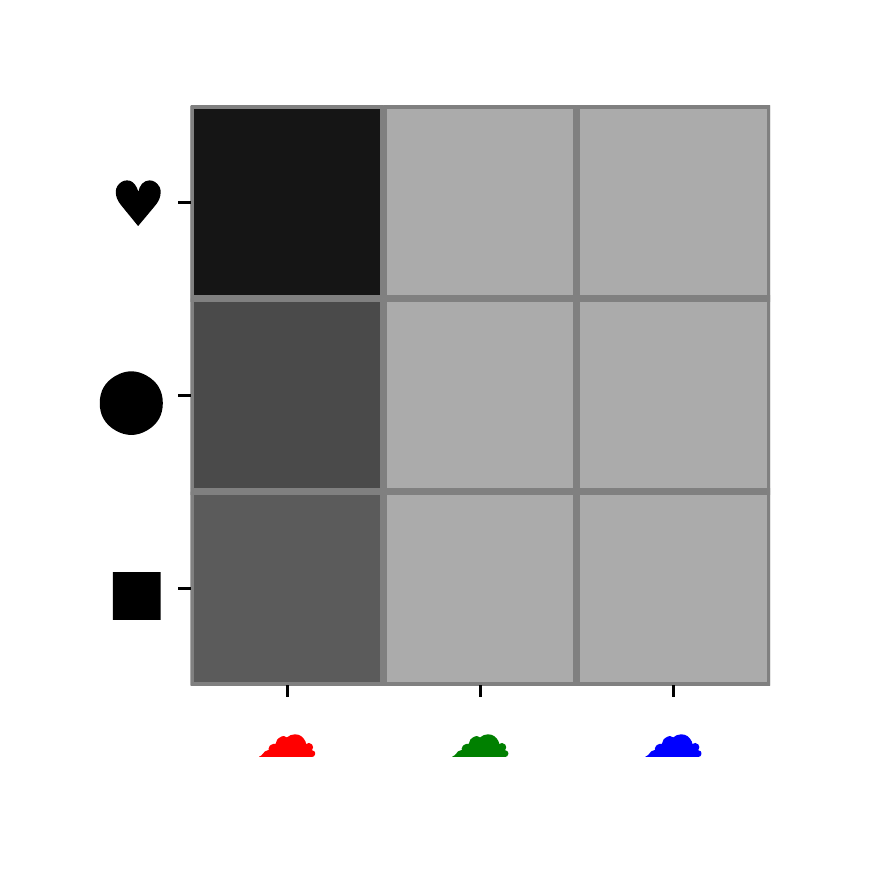}
    \caption{{\bf $\ptrain$: LDD.}}
    \label{fig:shift:LDD}
    \end{subfigure}
    \begin{subfigure}[b]{0.24\linewidth}
    \includegraphics[width=\linewidth]{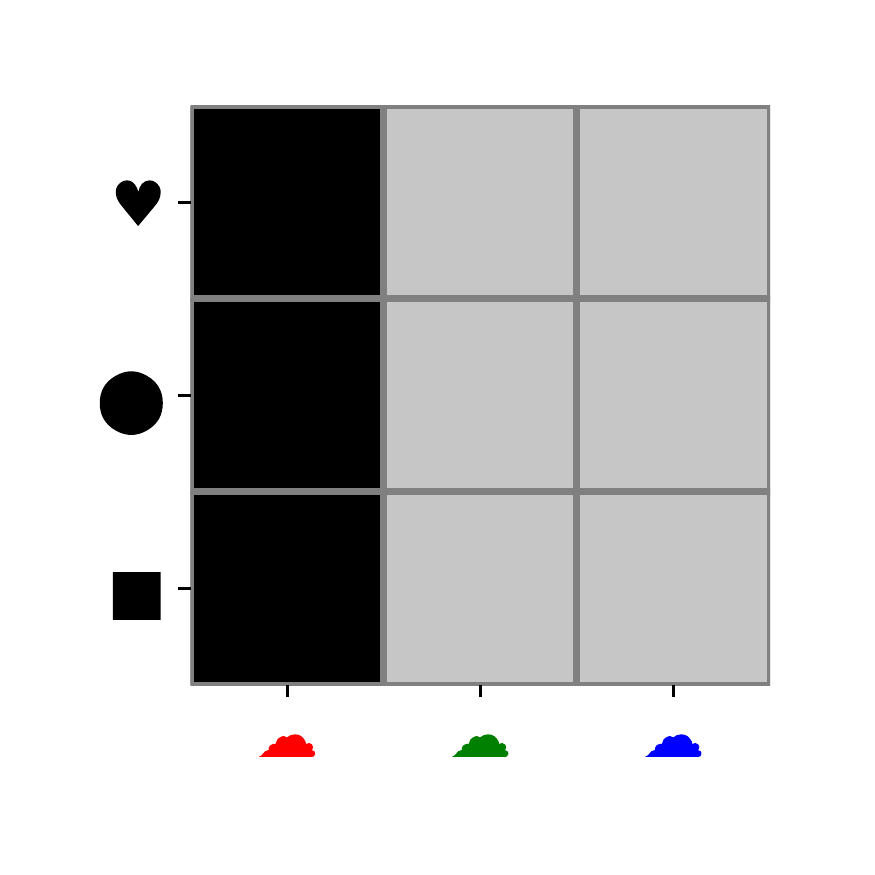}
    \caption{{\bf $\ptrain$: UDS.}}
    \label{fig:shift:SG}
    \end{subfigure}
    \begin{subfigure}[b]{0.24\linewidth}
    \includegraphics[width=\linewidth]{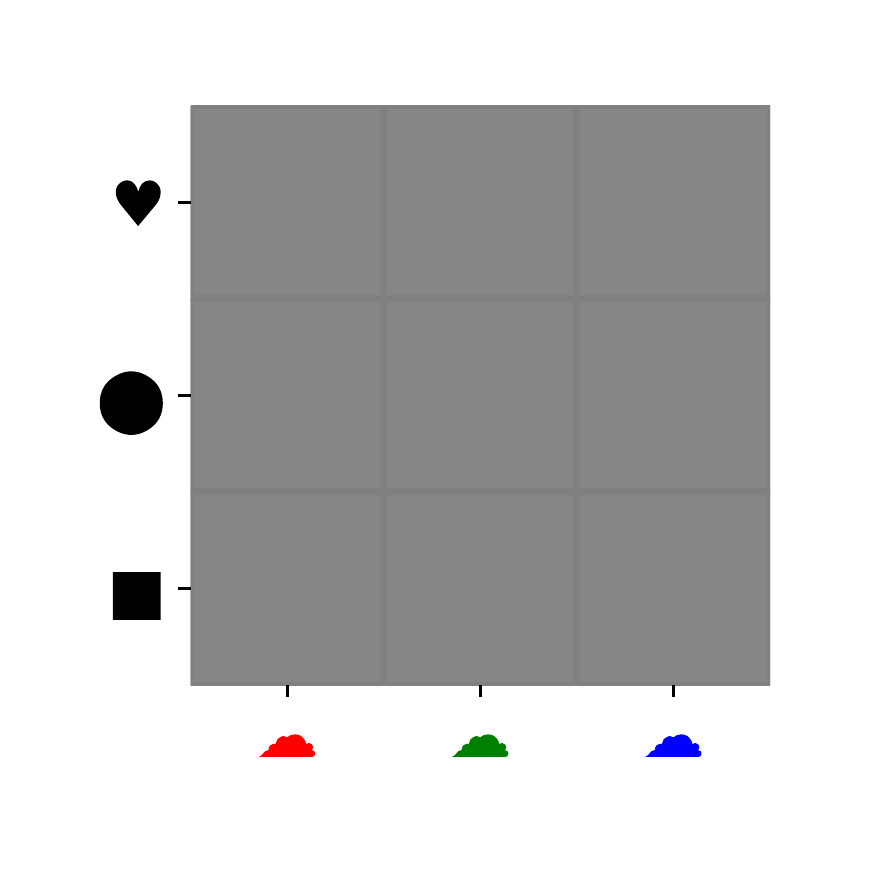}
    \caption{{\bf $\ptest$: $y^l, y^a$ are IID.}}
    \label{fig:shift:IID}
    \end{subfigure}
    \caption{Visualization of the joint distribution for the different shifts we consider on the \dsprites example.
    The lighter the color, the more likely the given sample. \figref{fig:shift:SC}-\ref{fig:shift:SG} visualise different shifts over $\ptrain(\attribute{l}, \attribute{a})$ discussed in \ref{sec:shifts}: {\em spurious correlation} (SC), {\em low-data drift} (LDD), and {\em unseen data shift} (UDS).
    \figref{fig:shift:IID} visualises the test set, where the attributes are uniformly distributed.
    }
    \label{fig:distribuionshift}
\end{figure}

While generalization in the above sense is desirable for machine learning models, it is not clear {\em why} a model $f$ trained on data from $\ptrain$ {\em should} generalise to $\ptest$.
It is worth noting that while $\ptrain$ and $\ptest$ can be different, they are both related to the true distribution $\ptrue$.
We take inspiration from disentanglement literature to express this relationship.
In particular, that we can view data as being decomposed into an underlying set of factors of variations.
\begin{comment}
Note that, unlike work in disentanglement literature, our aim is {\em not} to disentangle the data into these factors but to use this notion to build an understanding of how a model can learn to be robust to a given attribute and thereby generalise to unseen data.
The core hypothesis is that {\em if} a model can be built to learn invariances to a given attribute using data samples with different values for that attribute in $\ptrain$, then it can generalise to different distributions over that attribute in $\ptest$.
We assume images are composed of $\NumOfAttributes$ different, independent attributes $\attribute{1}, \attribute{2}, \dots, \attribute{\NumOfAttributes}$, each $\attribute{i}$ taking a discrete set of values, defined by the set $\attributeSet{i}$. As a running example, we use the color \dsprites dataset \citep{Matthey17}. 
We assume that $\attribute{1}$ defines the color with $\attributeSet{1} = \{\textrm{red}, \textrm{green}, \textrm{blue}\}$, and $\attribute{2}$ defines the shape with $\attributeSet{2} = \{ \textrm{ellipse}, \textrm{heart}, \textrm{square}\}$. 
Note that, in more realistic domains, the number of attributes may grow exponentially and even  potentially be uncountable.
In the classification setting, one attribute $y^l$ is designated the label (with $L = |\attributeSet{l}|$ possible values), and it is this attribute we wish $f$ to predict.
The aim is to be as invariant as possible to the other attributes.
\end{comment}
We formalise various distribution shifts using a latent variable model for the true data generation process:
% We will assume the following latent variable model for the true distribution $\ptrue$.
%There are some set of observations 
\begin{align}
\lat & \sim  \ptrue(\lat) & 
\attribute{i} & \sim  \ptrue(\attribute{i} | \lat)  \quad
i  =  1 \dots \NumOfAttributes & 
\vx & \sim  \ptrue(\vx | \lat) 
\end{align}
where $\lat$ denotes latent factors.
By a simple refactorization, we can write 
$$
p(\attribute{1:\NumOfAttributes}, \vx) = \ptrue(\attribute{1:\NumOfAttributes}) \int \ptrue(\vx|\lat) p(\lat|\attribute{1:\NumOfAttributes}) d\lat = \ptrue(\attribute{1:\NumOfAttributes}) p(\vx|\attribute{1:\NumOfAttributes}).
$$
Thus, the true distribution can be expressed as the product of the marginal distribution of the attributes with a conditional generative model.  
We assume that distribution shifts arise 
when a new marginal distribution for the attributes is chosen, such as $p(\attribute{1:\NumOfAttributes}) \neq \ptrain(\attribute{1:\NumOfAttributes}) \neq \ptest(\attribute{1:\NumOfAttributes})$, but otherwise
the conditional generative model is shared across all distributions, i.e., we have $\ptest(\attribute{1:\NumOfAttributes}, \vx) = \ptest(\attribute{1:\NumOfAttributes}) \int p(\vx|\lat) p(\lat|\attribute{1:\NumOfAttributes}) d\lat$, and similarly for $\ptrain$.

\begin{comment}
We focus on common cases of distribution shifts visualised in \figref{fig:distribuionshift}; we discuss these in more detail in section~\ref{sec:shifts}.
As a running example, we use the color \dsprites dataset \citep{Matthey17}; where in our notation $\attribute{1}$ defines the color with $\attributeSet{1} = \{\textrm{red}, \textrm{green}, \textrm{blue}\}$, and $\attribute{2}$ defines the shape with $\attributeSet{2} = \{ \textrm{ellipse}, \textrm{heart}, \textrm{square}\}$. 
To generate a dataset $\textrm{e} \in \{\textrm{train}, \textrm{test}, \textrm{true}\}$, 
we imagine that the data collector (intentionally or implicitly) 
selects some marginal distribution over attributes $p_\textrm{e}(\attribute{1:\NumOfAttributes})$, for example they select mostly blue ellipses and red hearts. In practice, this can occur due to selection bias during data creation and labelling process and
induces a new joint distribution over latent factors and attributes  $p_\textrm{e}(\lat, \attribute{1:\NumOfAttributes}) = \ptrue(\lat| \attribute{1:\NumOfAttributes}) p_\textrm{e}(\attribute{1:\NumOfAttributes})$.
Consequently, we get images with a different joint distribution $p_\textrm{e}(\vx, \attribute{1:\NumOfAttributes}) = \int \ptrue(\vx|\lat) p_\textrm{e}(\lat, \attribute{1:\NumOfAttributes})$.
\end{comment}

% \sven{Here's my take on the above paragraph:}
To provide more context, as a running example, we use the color \dsprites dataset \citep{Matthey17}; where in our notation $\attribute{1}$ defines the color with $\attributeSet{1} = \{\textrm{red}, \textrm{green}, \textrm{blue}\}$, and $\attribute{2}$ defines the shape with $\attributeSet{2} = \{ \textrm{ellipse}, \textrm{heart}, \textrm{square}\}$.
We can imagine that a data collector (intentionally or implicitly) selects some marginal distribution over attributes $\ptrain(\attribute{1:\NumOfAttributes})$ when training; for example they select mostly blue ellipses and red hearts.
This induces a new joint distribution over latent factors and attributes: $\ptrain(\lat, \attribute{1:\NumOfAttributes}) = \ptrue(\lat| \attribute{1:\NumOfAttributes}) \ptrain(\attribute{1:\NumOfAttributes})$.
Consequently, during training, we get images with a different joint distribution $\ptrain(\vx, \attribute{1:\NumOfAttributes}) = \int \ptrue(\vx|\lat) \ptrain(\lat, \attribute{1:\NumOfAttributes})$.
This similarly applies when collecting data for the test distribution.
We focus on common cases of distribution shifts visualized in \figref{fig:distribuionshift}; we discuss these in more detail in section~\ref{sec:shifts}.

\begin{comment}
Intuitively, the above model assumes the presence of latent `features' $\lat$, 
where conditional on these features, the observed image and observable attributes
are conditionally independent. 
Some, unknown generative model, samples $\lat$ and generates the image $\vx$ and corresponding attributes $y^1\dots y^k$. This is the data we observe.
\end{comment}

%In this model, distribution shifts arise when the marginal distribution of attributes varies between %$\ptrain$ and $\ptest$. 

The goal of enforcing robustness to distribution shifts is to maintain performance when the data generating distribution $\ptrain$ changes.
In other words, we would like to minimize risk on $\ptrue, \ptest$ given {\em only} access to $\ptrain$.
We can achieve robustness in the following ways:
\begin{itemize}
\item {\bf Weighted resampling.} We can resample the training set using importance weights $W(\attribute{1:\NumOfAttributes}) = p(\attribute{1:\NumOfAttributes}) / \ptrain(\attribute{1:\NumOfAttributes})$.
This means that given the attributes, the $i$-th data point $(\attribute{1:\NumOfAttributes}_i, \vx_i)$ in the training set is used with probability  $W(\attribute{1:\NumOfAttributes}_i) / \sum_{i'=1}^n W(\attribute{1:\NumOfAttributes}_{i'})$ rather than $1/n$. We refer to this empirical distribution as $p_\textrm{reweight}$.
This procedure requires access to the true distribution of attributes $p(\attribute{1:\NumOfAttributes})$, so to avoid bias and improve fairness, it is often assumed that all combinations of attributes happen uniformly at random.
\item {\bf Data Augmentation}: 
Alternatively, we can learn a generative model $\hat{p}(\vx| \attribute{1:\NumOfAttributes})$ from the training data that aims to approximate $\int \ptrue(\vx| \lat) \ptrue(\lat | \attribute{1:\NumOfAttributes}) dz$, as the true conditional generator is by our assumption the same over all (e.g.~train and test) distributions.
If such a conditional generative model can be learned, we can sample new synthetic data at training time (e.g.~according to the true distribution $\ptrue(\attribute{1:\NumOfAttributes})$) to correct for the distribution shift.
More precisely,  we can generate data from the augmented distribution $p_\textrm{aug} = (1 - \alpha) p_\textrm{reweight} + \alpha \hat{p}(\vx| \attribute{1:\NumOfAttributes}) \ptrue(\attribute{1:\NumOfAttributes})$ and train a supervised classifier on this augmented dataset.
Here, $\alpha \in [0, 1]$ is the percentage of synthetic data used for training.
\item {\bf Representation Learning}: An alternative factorization of a data generating distribution (e.g.~train) is 
$p_\textrm{train}(\attribute{1:\NumOfAttributes}, \vx) = \int \ptrue(\lat | \vx) p_\textrm{train}(\attribute{1:\NumOfAttributes}| \lat) d\lat$.
We can learn an unsupervised representation that approximates $\ptrue(\lat | \vx)$  using the training data only, and attach a classifier to learn a task specific head that approximates $p_\textrm{train}(\attribute{l}| \lat)$. Again, by our assumption $\ptrue(\lat | \vx) \propto \ptrue(\vx|\lat) \ptrue(\lat)$. Given a good guess of the true prior, the learned representation would not be impacted by the specific attribute distribution and so generalise to $\ptest,\ptrue$.
\end{itemize}

\subsection{Distribution Shifts}
\label{sec:shifts}
% Intuitively, certain shifts may matter more than others. 
While distribution shifts can happen in a continuum, we consider three types of shifts, inspired by real-world challenges. 
We discuss these shifts and two additional, real-world inspired conditions.

\paragraph{Test distribution $\ptest$.}
% \olivia{Something to link invariance of attributes to doing well on this objective?}
We assume that the attributes are distributed uniformly: 
 $\ptest(y^{1:K}) = 1 / \prod_i |\attributeSet{i}|$.
This is desirable, as all attributes are represented and a-priori independent.
% If we believe that the ideal classifier over $\ptest$ should classify images independently of a given attribute $y^a$; this can be achieved by learning a classifier on $\ptrain$ that learns to ignore $y^a$, potentially
% generalising to unseen values of the set $\attributeSet{a}$.

\begin{figure}
    \begin{subfigure}[b]{0.3\linewidth} 
    \footnotesize

    \includegraphics[trim={0 0 10.1cm 0cm},clip,width=\linewidth]{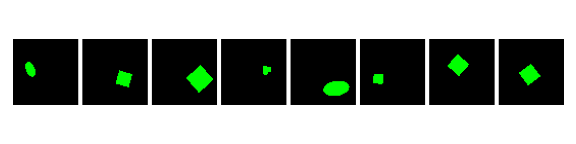}
    
    \vspace{-3.5em}
    
    \includegraphics[trim={0 0 10.1cm 0cm},clip,width=\linewidth]{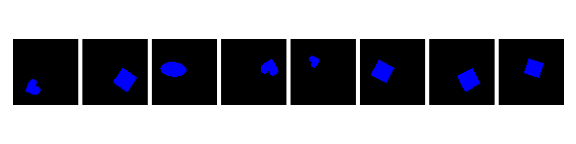}
    
    \vspace{-1.7em}
    \caption*{\dsprites.}
    \end{subfigure}
    \begin{subfigure}[b]{0.3\linewidth} 
    \footnotesize

    \includegraphics[trim={0 0 10.1cm 0cm},clip,width=\linewidth]{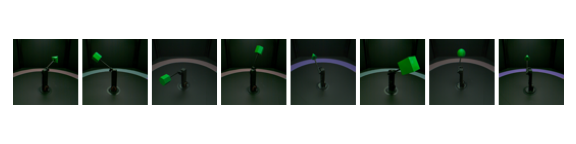}
    
    \vspace{-3.5em}
    
    \includegraphics[trim={0 0 10.1cm 0cm},clip,width=\linewidth]{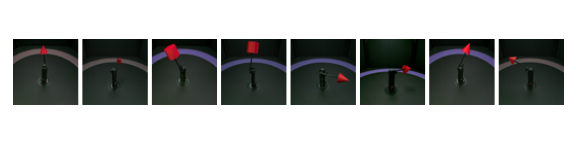}
    
    \vspace{-1.7em}
    \caption*{\mpithreed.}
    \end{subfigure}
    \begin{subfigure}[b]{0.3\linewidth} 
    \footnotesize

    \includegraphics[trim={0 0 10.1cm 0cm},clip,width=\linewidth]{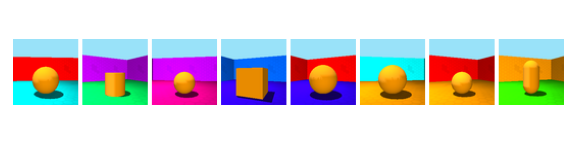}
    
    \vspace{-3.5em}
    
    \includegraphics[trim={0 0 10.1cm 0cm},clip,width=\linewidth]{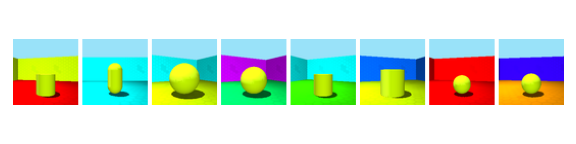}
    
    \vspace{-1.7em}
    \caption*{\shapes.}
    \end{subfigure}
    \begin{subfigure}[b]{0.3\linewidth} 
    \footnotesize
    \includegraphics[trim={0 0 10.1cm 0cm},clip,width=\linewidth]{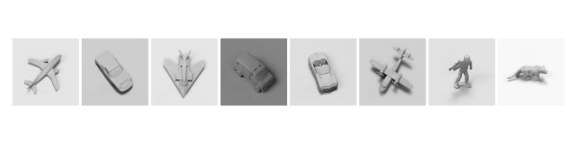}
    
    \vspace{-3.5em}

    \includegraphics[trim={0 0 10.1cm 0cm},clip,width=\linewidth]{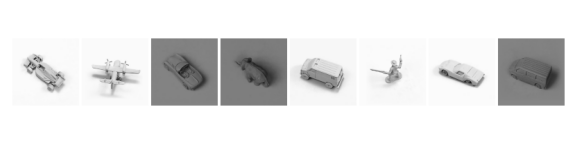}
    
    \vspace{-1.7em}
    \caption*{\smallnorb.}
    \end{subfigure}
    \begin{subfigure}[b]{0.3\linewidth} 
    \footnotesize

    \includegraphics[trim={0 0 10.1cm 0cm},clip,width=\linewidth]{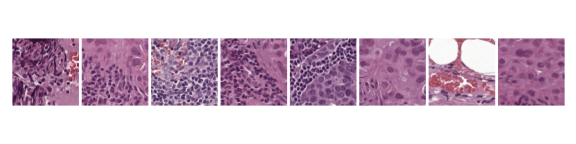}
    
    \vspace{-3.5em}
    
    \includegraphics[trim={0 0 10.1cm 0cm},clip,width=\linewidth]{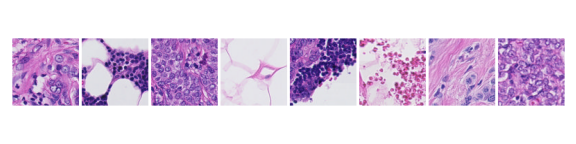}
    
    \vspace{-1.7em}
    \caption*{\camelyon.}
    \end{subfigure}
    \begin{subfigure}[b]{0.3\linewidth} 
    \footnotesize
    \includegraphics[trim={0 0 10.1cm 0cm},clip,width=\linewidth]{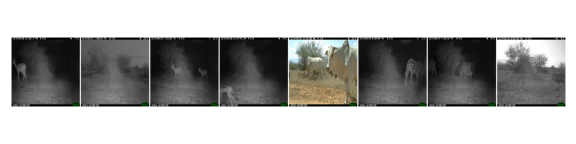}
    
    \vspace{-3.5em}
    
    \includegraphics[trim={0 0 10.1cm 0cm},clip,width=\linewidth]{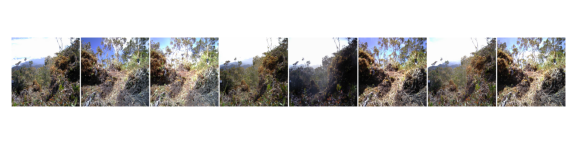}

    \vspace{-1.7em}
    \caption*{\iwildcam.}
    \end{subfigure}
    \caption{{\bf Dataset samples}. Each row fixes an attribute (e.g.~color for \dsprites, \mpithreed, \shapes; azimuth for \smallnorb; hospital for \camelyon; and location for \iwildcam).}
    \label{fig:datasetvisualisation}
\end{figure}

\paragraph{Shift 1: Spurious correlation -- Attributes are correlated under $\ptrain$ but not $\ptest$.} 
Spurious correlation arises in the wild for a number of reasons including capture bias, environmental factors, and geographical bias \citep{Beery18,Torralba11}.
% Capture bias arises when data is captured in a consistent manner (for example, mugs are photographed with their handle to the right) \citep{Torralba11}.
% Environmental factors create bias; for example, certain animals are often seen in the same terrain (camels in the desert), leading to a spurious correlation between the animal and the background \citep{Beery18}.
% Finally, geographical bias can occur when certain languages are used to collect data (e.g.~querying {\em wedding} in English as opposed to in Hindi will give very different samples of what a wedding is).
These spurious correlations lead to surprising results and poor generalization. Therefore, it is important to be able to build models that are robust to such challenges.
In our framework, spurious correlation arises when two attributes $y^a$, $y^b$ are correlated at training time, but this is not true of  $\ptest$, for which attributes are independent: $\ptrain(y^a|y^1\dots y^b \dots y^K) > \ptrain(y^a|y^1\dots y^{b-1},y^{b+1} \dots y^K)$.
This is especially problematic when one attribute $y^b$ is $y^l$, the label. 
% If $y^a$ is easier to predict, the model may rely on the wrong attribute, leading to poor generalization performance.
Using the running \dsprites example, shape and color may be correlated and the model may find it easier to predict color.
% If the model finds it easier to predict color, it may rely on it for shape classification. 
If color is the label, the model will generalise well. However, if the aim is to predict shape, the model's reliance on color will lead to poor generalization.

\paragraph{Shift 2: Low-data drift -- Attribute values are unevenly distributed under $\ptrain$ but not under $\ptest$.}
Low-data drift arises in the wild (e.g.~in \citep{Buolamwini18} for different ethnicities) when data has not been collected uniformly across different attributes.
% For example, \cite{Buolamwini18} demonstrated that if data comes mostly from one gender or ethnicity, this leads to inconsistent and unfair performance across different genders and ethnicities.
When deploying models in the wild, it is important to be able to reason and have confidence that the final predictions will be consistent and fair across different attributes.
In the framework above, low-data shifts arise when certain values in the set $\sA^a$ of an attribute $y^a$ are sampled with a much smaller probability than in $\ptest$: $\ptrain(y^a = v) \ll \ptest(y^a = v)$.
Using the \dsprites example, only a handful of red shapes may be seen at training time, yet in $\ptest$ all colors are sampled with equal probability.

\paragraph{Shift 3: Unseen data shift -- Some attribute values are unseen under $\ptrain$ but are under $\ptest$.}
This is a special case of {\em shift 2: low-data drift} which we make explicit due to its important real world applications.
Unseen data shift arises when a model, trained in one setting is expected to work in another, disjoint setting.
For example:  a model trained to classify animals on images at certain times of day should generalise to other times of day.
In our framework, {\em unseen data shift} arises when some values in the set $\sA^a$ of an attribute $y^a$ are unseen in $\ptrain$ but are in $\ptest$:
\begin{align}
    \ptrain(y^a = v) = 0 \quad \quad \quad
    \ptest(y^a = v) > 0 \quad \quad \quad
    \left| \{ v | \ptrain(y^a = v)\} \right| > 1
\end{align}
This is a stronger constraint than in standard out-of-distribution generalization (see \secref{sec:relatedwork}), as multiple values for $\sA^a$ must be seen under $\ptrain$, which allows the model to learn invariance to $y^a$.
In the \dsprites example, the color red may be unseen at train time but all colors are in $\ptest$.

% This setup is related to OOD generalization literature (reviewed in \secref{sec:relatedwork}); however, we constrain the types of shifts being considered. 
% Instead of only imposing that class labels must be the same between $\ptrain$ and $\ptest$, we impose a stronger constraint: that an attribute value may change, but that multiple values of this attribute must have been seen in $\ptrain$.
% Unlike in a standard OOD generalization dataset, this stronger constraint gives a clear notion of why a model should be able to generalise: it should be able to learn invariance to the given attribute, as it has seen multiple samples from it and different values for it.

\paragraph{Discussion.}
We choose these sets of shifts as they are the building blocks of more complex distribution shifts.
Consider the simplest case of two attributes: the label and a nuisance attribute.
If we consider the marginal distribution of the label, it decomposes into two terms: the conditional probability and the probability of a given attribute value: $p(y^l) = \sum_{y^a} p(y^l|y^a) p(y^a)$.
The three shifts we consider control these terms independently: {\em unseen data shift} and {\em low-data drift} control $p(y^a)$ whereas {\em spurious correlation} controls $p(y^l|y^a)$.
The composition of these terms describes any distribution shift for these two variables.

\subsection{Conditions}
% We additionally evaluate these approaches under two conditions, inspired by real world challenges.

{\bf Label noise.}
We investigate the change in performance due to noisy information. 
This can arise when there are disagreements and errors among the labellers (e.g.~in medical imaging \citep{Castro20}).
We model this as an observed attribute (e.g.~the label) being corrupted by noise.
$\hat{y}^i \sim c(y^i)$, where ${y}^i \in \attributeSet{i}$ is the true label, $\hat{y}^i \in \attributeSet{i}$ the corrupted, observed one, and $c$ the corrupting function.

{\bf Dataset size.}
We investigate how performance changes with the size of the training dataset.
This setting arises when it is unrealistic or expensive to collect additional data (e.g.~in medical imaging or in camera trap imagery).
Therefore, it is important to understand how performance degrades given fewer total samples.
We do this by limiting the total number of  samples  from $\ptrain$.

\begin{figure}
\begin{floatrow}
\ffigbox{%
  \centering
  \includegraphics[width=\linewidth,trim={3.5cm .5cm 4.2cm 2cm},clip]{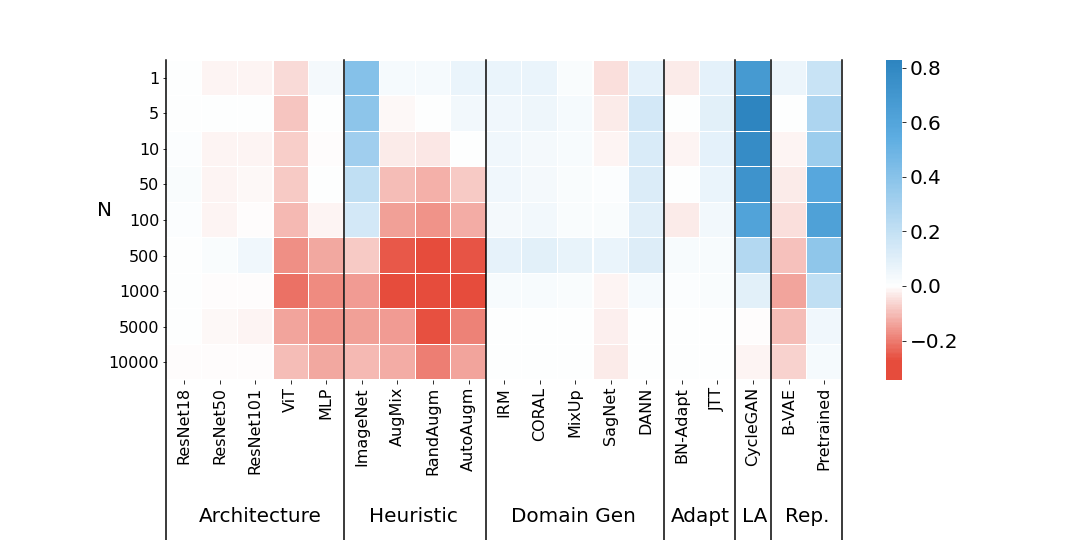}
}{%
  \caption{{\bf Spurious Correlation.} We use all correlated samples and vary the number of samples $N$ from the true, uncorrelated distribution. We plot the percentage change over the baseline ResNet, averaged over all seeds and datasets. Blue is better, red worse. \cyclegan performs consistently best while ImageNet augmentation and pretraining on ImageNet also consistently boosts performance. \label{fig:spuriouscorrelation_hm}}%
}
\ffigbox{%
  \includegraphics[width=\linewidth,trim={3.5cm .5cm 4.2cm 2cm},clip]{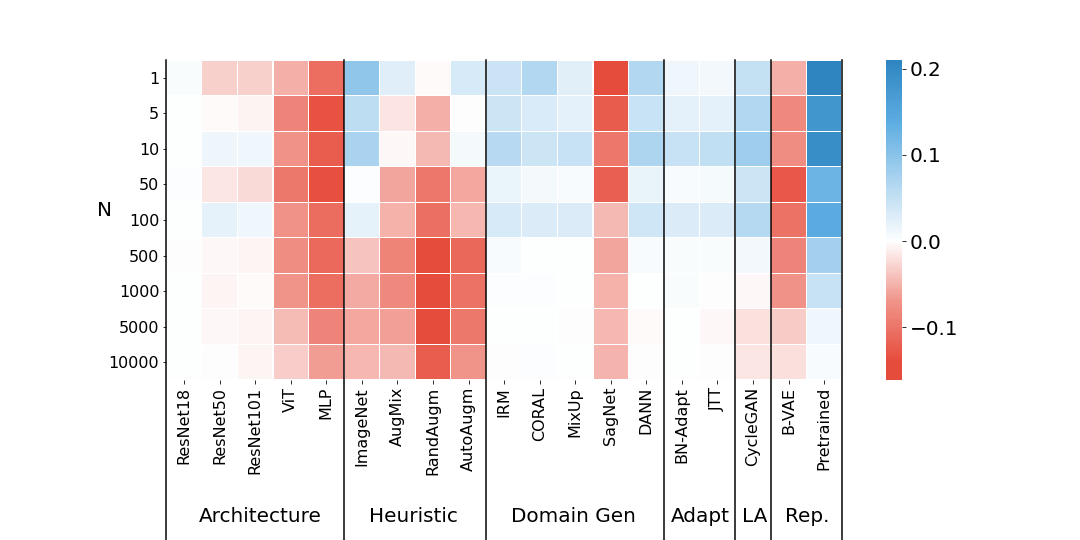}
}{%
  \caption{{\bf Low-data drift.} We use all samples from the high data regions and vary the number of samples $N$ from the low-data region. We plot the percentage change over the baseline ResNet, averaged over all seeds and datasets. Blue is better, red worse. Pretraining on ImageNet performs consistently best, while \cyclegan, most domain generalization methods and ImageNet augmentation also provide some boost in performance. \label{fig:lowdata_hm}}%
}
\end{floatrow}

\end{figure}

\section{Models evaluated}
\label{sec:modelsetup}

We evaluate 19 algorithms to cover a broad range of approaches that can be used to improve model robustness to distribution shifts and demonstrate how they relate to the three ways to achieve robustness, outlined in \secref{sec:robustnessframework}.
We believe this is the first paper to comprehensively evaluate a large set of different approaches in a variety of settings.
These algorithms cover the following areas: architecture choice, data augmentation, domain adaptation, adaptive approaches and representation learning.
Further discussion on how these models relate to our robustness framework is in \appref{sec:app:methods}.

{\bf Architecture choice.} We evaluate the following standard vision models: ResNet18, ResNet50, ResNet101 \citep{He16}, ViT \citep{Dosovitskiy20}, and an MLP \citep{Vapnik92}. We use weighted resampling $p_\textrm{reweight}$ to oversample from the parts of the distribution that have a lower probability of being sampled from under $\ptrain$. 
Performance depends on how robust the learned representation is to distribution shift.

 {\bf Heuristic data augmentation.} These approaches attempt to approximate the true underlying generative model $p(\vx|y^{1:K})$ in order to improve robustness. We analyze the following augmentation methods: standard ImageNet augmentation \citep{He16}, AugMix without JSD \citep{Hendrycks20b}, RandAugment \citep{Cubuk20}, and AutoAugment \citep{Cubuk19}. 
Performance depends on how well the heuristic augmentations approximate the true generative model.

{\bf Learned data augmentation.} These approaches approximate the true underlying generative model $p(\vx|y^{1:K})$ by learning augmentations conditioned on the nuisance attribute. The learned augmentations can be used to transform any image $\vx$ to have a new attribute, while keeping the other attributes fixed. We follow \cite{Goel20}, who use  \cyclegan \citep{Zhu17}, but we do not use their SGDRO objective in order to evaluate the performance of learned data augmentation alone. 
Performance depends on how well the learned augmentations approximate the true generative model.

{\bf Domain generalization.} These approaches aim to recover a representation $z$ that is independent of the attribute: $p(y^a, z) = p(y^a)p(z)$ to allow generalization over that attribute. We evaluate IRM \citep{Arjovsky18}, DeepCORAL \citep{Sun16}, domain MixUp \citep{Gulrajani20}, DANN \citep{Ganin16}, and SagNet \citep{Nam21}. 
Performance depends on the invariance of the learned representation $z$.

{\bf Adaptive approaches.} These works modify $p_\textrm{
    reweight}$ dynamically. We evaluate JTT \citep{Liu21} and BN-Adapt \citep{Schneider20}. 
    These methods do not give performance guarantees.

{\bf Representation learning. } These works aim to learn a robust representation of $z$ that describes the true prior. We evaluate using a $\beta$-VAE \citep{Higgins16} and pretraining on ImageNet \citep{Deng09}. 
Performance depends on the quality of the learned representation for the specific task.

\section{Experiments}
We first introduce the datasets and experimental setup.
We evaluate the 19 different methods across these six datasets, three distribution shifts, varying label noise, and dataset size.
We plot aggregate results in \multfigref{fig:spuriouscorrelation_hm}{fig:fixeddata_hm} and complete results in the appendix in  \multfigref{fig:stronggeneralization}{fig:lowdatadrift}.
We discuss the results by distilling them into seven concrete takeaways in \secref{sec:exp:takeaways} and four practical tips in \secref{sec:practicaltips}.

\paragraph{Datasets.}
We evaluate these approaches on six vision, classification datasets -- \dsprites \citep{Matthey17}, \mpithreed \citep{Gondal19}, \smallnorb \citep{LeCun04}, \shapes \citep{Burgess18b}, \camelyon \citep{Wilds2020,Bandi18}, and \iwildcam \citep{Wilds2020,Beery18}. 
These datasets consist of multiple (potentially an arbitrarily large number) attributes.
We select two attributes $y^l, y^a$ for each dataset and make one $y^l$ the label.
We then use these two attributes to build the three shifts.
Visualizations of samples from the datasets are given in \figref{fig:datasetvisualisation} and further description in \appref{sec:app:datasetdetails}. 
We discuss precisely how we set up the shifts, choose the attributes, and additional conditions for these datasets in \appref{sec:app:datasetshifts}.

\begin{figure}
\floatbox[{\capbeside\thisfloatsetup{capbesideposition={right,center},capbesidewidth=0.5\linewidth}}]{figure}[\FBwidth]
{
\caption{{\bf Unseen data shift.} We rank the methods (where best is $1$, worst $19$) for each dataset and seed and plot the rankings, with the overall median rank as the black bar. Pretraining on ImageNet and ImageNet augmentation perform consistently best. DANN, CycleGAN and other heuristic augmentations perform consistently well.\label{fig:stronggeneralization_hm}}%
}
{\includegraphics[width=\linewidth]{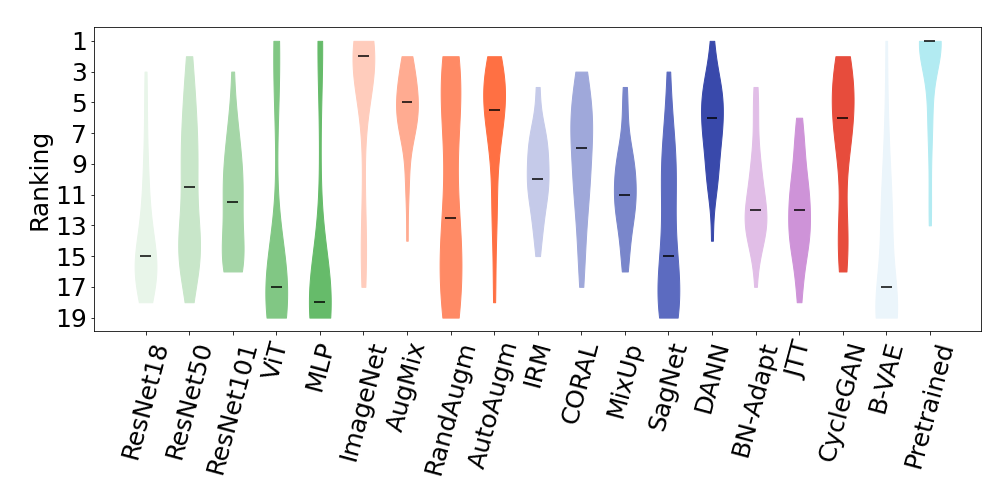}}
\vspace{-1.5em}
\end{figure}

% \subsection{Model selection and randomness of runs.}
\paragraph{Model selection.}
When investigating heuristic data augmentation, domain generalization, learned augmentation, adaptive approaches, and representation learning, we use a ResNet18 for the simpler, synthetic datasets (\dsprites, \mpithreed, \shapes, and \smallnorb) but a ResNet50 for the more complex, real world ones (\camelyon and \iwildcam). 
To perform model selection, we choose the best model according to the in-distribution validation set ($\valid \subset \train$).
In the {\em unseen data shift} setting for the \camelyon and \iwildcam, we use the given out-of-distribution validation set, which is a different, distinct set in $\dataset$ that is independent of $\train, \test$.
(We consider using the in-distribution validation set in \appref{sec:app:results:idvsoood}.)

\paragraph{Hyperparameter choices.}
We perform a sweep over the hyperparameters (the precise sweeps are given in \appref{sec:app:sweeps}).
We run each set of hyperparameters for five seeds for each setting.
To choose the best model for each seed, we perform model selection over {\em all} hyperparameters using the top-1 accuracy on the validation set.
In the {\em low-data} and {\em spurious correlation} settings, we choose a different set of samples from the low-data region with each seed.
We report the mean and standard deviation over the five seeds.

\subsection{Takeaways}
\label{sec:exp:takeaways}

{\bf Takeaway 1: While we can improve over ERM, no one method always performs best.}
% While it is tempting to claim that a model is state-of-the-art based on a given dataset, our results show that such claims are misleading, as t
The relative performance between methods varies across datasets and shifts.
 Under {\em spurious correlation} (\figref{fig:spuriouscorrelation_hm}), \cyclegan consistently performs best but in \figref{fig:lowdata_hm}, under {\em low-data drift}, pretraining consistently performs best.
Under {\em unseen data shift} (\figref{fig:stronggeneralization_hm}), pretraining is again one of the best performing models.
However, if we drill down on the results in \figref{fig:stronggeneralization} (\appref{sec:app:results:complete}), we can see pretraining performs best on the synthetic datasets, but not on \camelyon (where using augmentation or DANN is best) or \iwildcam (where using ViT or an MLP is best).

{\bf Takeaway 2: Pretraining is a powerful tool across different shifts and datasets.}
While pretraining is not always helpful (e.g.~in \appref{sec:app:results:complete} on \camelyon in \multfigref{fig:stronggeneralization}{fig:spuriouscorrelation}, \iwildcam in \multfigref{fig:stronggeneralization}{fig:spuriouscorrelation}), it often provides a strong boost in performance.
This is presumably because the representation $z$ learned during pretraining is helpful for the downstream task.
For example, the representation may have been trained to be invariant to certain useful properties (e.g.~scale, shift, and color). 
If these properties are useful on the downstream tasks, then the learned representation should improve generalization.

{\bf Takeaway 3: Heuristic augmentation improves generalization {\em if} the augmentation describes an attribute.}
In all settings (\multfigref{fig:spuriouscorrelation_hm}{fig:stronggeneralization_hm}), ImageNet augmentation generally improves performance.
However, RandAugment, AugMix, and AutoAugment have more variable performance (as further shown in \multfigref{fig:stronggeneralization}{fig:lowdatadrift}).
These methods are compositions of different augmentations.
We investigate the impact of each augmentation in RandAugment in \appref{sec:app:results:randaugment} and find variable performance.
Augmentations that approximate the true underlying generative model $p(\vx|y^{1:K})$ lead to the best results; otherwise, the model may waste capacity.
For example, on \camelyon (which consists of cell images), color jitter harms performance but on \shapes and \mpithreed it is essential.

{\bf Takeaway 4: Learned data augmentation is effective across different conditions and distribution shifts.}
% An uncommon approach in practice, but learned data augmentation is surprisingly effective.
% This approach learns augmentations automatically based on auxiliary attribute information.
% If the generative model is perfect, the samples from the learned augmentation are by design samples from the true generative model.
This approach is highly effective in the {\em spurious correlation} setting (\figref{fig:spuriouscorrelation_hm}). It can also help in the {\em low-data} and {\em unseen data shift} settings (\figref{fig:lowdata_hm},\ref{fig:stronggeneralization_hm}) (though the gains for these two shifts are not as large as for pretraining).
The effectiveness of this approach can be explained by the fact that if the augmentations are learned perfectly, then augmented samples by design are from the true underlying generative model and can cover missing parts of the distribution.

{\bf Takeaway 5: Domain generalization algorithms offer limited performance improvement.}
In some cases these methods (in particular DANN) do improve performance, most notably in the {\em low-data drift} and {\em unseen data shift} settings (\multfigref{fig:lowdata_hm}{fig:stronggeneralization_hm}).
However, this depends on the dataset (see \multfigref{fig:stronggeneralization}{fig:lowdatadrift}) and performance is rarely much better than using heuristic augmentation.

\begin{figure}
\begin{floatrow}
\ffigbox{
\centering
\includegraphics[width=\linewidth,trim={2cm 0cm 4.5cm 2cm},clip]{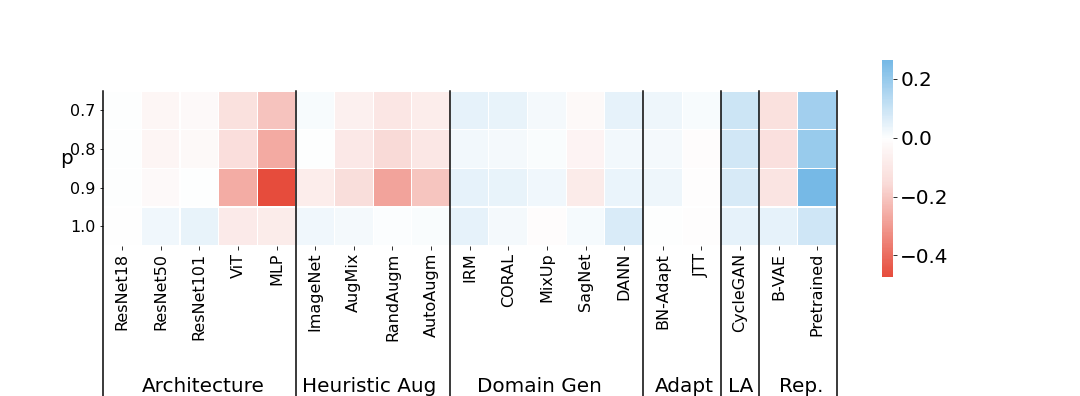}
}{
    \caption{{\bf Condition 1: Noisy labels.} We vary the amount of noise $p$ in the labels. We plot the percentage change over the baseline ResNet, averaged over all seeds and datasets. \label{fig:noise_hm}}
}
\ffigbox{
\centering
\includegraphics[width=\linewidth,trim={2cm 0cm 4.5cm 2cm},clip]{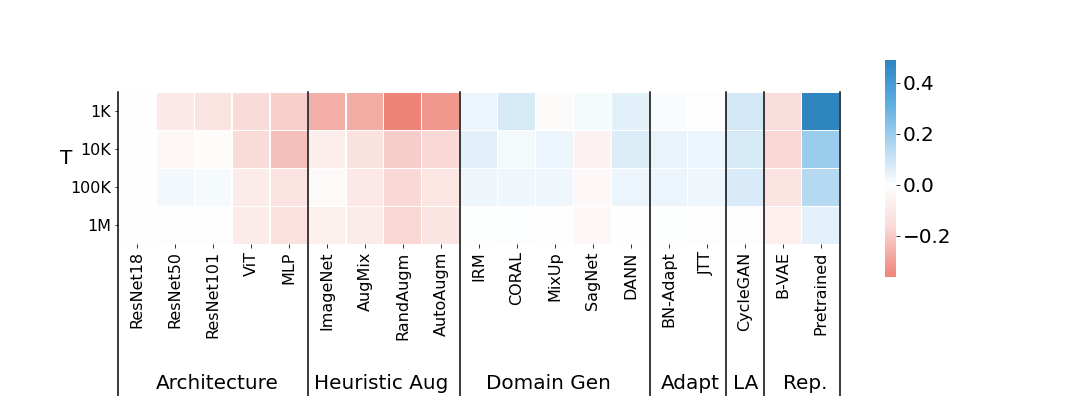}
}{
    \caption{{\bf Condition 2: Fixed data.} We vary the total size of the dataset $T$. We plot the percentage change over the baseline ResNet, averaged over all seeds and datasets. \label{fig:fixeddata_hm}}
}
\end{floatrow}
\end{figure}

{\bf Takeaway 6: The best algorithms may differ under the precise conditions.}
When labels have varying noise in \figref{fig:noise_hm}, relative performance is reasonably consistent.
When the dataset size decreases in \figref{fig:fixeddata_hm}, heuristic augmentation methods perform poorly.
However, using pretraining and learned augmentation is consistently robust.
% This demonstrates that model performance may vary under the precise conditions in which it is being used.

{\bf Takeaway 7: The precise attributes we consider directly impacts the results. }
For example, on \dsprites, if we make color $y^l$ and shape $y^a$, we find that {\em all} methods generalise perfectly in the {\em unseen data shift} setting (as demonstrated in \appref{sec:app:results:diffattributes}) unlike when shape is $y^l$ (\figref{fig:stronggeneralization}).

\subsection{Practical tips}
\label{sec:practicaltips}
While there is no free lunch in terms of the method to choose, we recommend the following tips.

{\bf Tip 1: If heuristic augmentations approximate part of the true underlying generative model, use them.}
% Heuristic augmentations help significantly if they approximate part of the true underlying generative model.
Under this constraint, heuristic augmentations can significantly improve performance; this should be a first point of call.
How to heuristically choose these augmentations without exhaustively trying all possible combinations is an open research question.

{\bf Tip 2: If heuristic augmentations do not help, learn the augmentation.}
If the true underlying generative model cannot be readily approximated with heuristic techniques, but some subset of the generative model can be learned by conditioning on known attributes, this is a promising way to further improve performance.
How to learn the underlying generative model directly from data and use this for augmentation is a promising area to explore more thoroughly.

{\bf Tip 3: Use pretraining.}
In general, pretraining was found to be a useful way to learn a robust representation.
While this was not true for all datasets (e.g.~\camelyon, \iwildcam), performance could be dramatically improved by pretraining (\dsprites, \mpithreed, \smallnorb, \shapes).
% Therefore, using pretraining on more data and preferably on data related to the downstream task will likely improve performance.
An area to be investigated is the utility of self-supervised pre-training.

{\bf Tip 4: More complex approaches lead to limited improvements.}
Domain generalization, adaptive approaches and disentangling lead to limited improvements, if any, across the different datasets and shifts.
Of these approaches, DANN performs generally best.
How to make these approaches generically useful for robustness is still an open research question.

\section{Discussion}
\label{sec:discussion}

Our experiments demonstrate that no one method performs best over all shifts and that performance is dependent on the precise attribute being considered.
This leads to the following considerations.

{\bf There is no way to decide a-priori on the best method given only the dataset.} It would be helpful for practitioners to be able to select the best approaches without requiring comprehensive evaluations and comparisons.
Moreover, it is unclear how to pinpoint the precise distribution shift (and thereby methods to explore) in a given application.
This should be an important future area of investigation.

{\bf We should focus on the cases where we have knowledge about the distribution shift.}
We found that the ability of a given algorithm to generalize depends heavily on the attribute and dataset being considered. Instead of trying to make one algorithm for any possible shift, it makes sense to have adaptable algorithms which can use auxiliary information if given.
Moreover, algorithms should be evaluated in the context for which we will use them.

{\bf It is pivotal to evaluate methods in a variety of conditions. }
Performance varies due to the number of examples, amount of noise, and size of the dataset.
Thus it is important to perform comprehensive evaluations when comparing different methods, as in our framework.
This gives others a more realistic view of different models' relative performance in practice.

\section{Related Work}
\label{sec:relatedwork}

We briefly summarize benchmarks on distribution shift, leaving a complete review to \appref{sec:app:relatedwork}.

\paragraph{Benchmarking robustness to out of distribution (OOD) generalization.}
While a multitude of methods exist that report improved OOD generalization, \cite{Gulrajani20} found that in actuality no evaluated method performed significantly better than a strong ERM baseline on a variety of datasets.
However, \cite{Hendrycks21} found that, when we focus on better augmentation, larger models and pretraining, we can get a sizeable boost in performance.
This can be seen on the \cite{Wilds2020} benchmark (the largest boosts come from larger models and better augmentation).
Our work is complementary to these methods, as we look at a range of approaches (pretraining, heuristic augmentation, learned augmentation, domain generalisation, adaptive, disentangled representations) on a range of both synthetic and real-world datasets.
Moreover, we allow for a fine-grained analysis of methods over different distribution shifts.

\paragraph{Benchmarking spurious correlation and low-data drift.}
 Studies on fairness and bias (surveyed by \cite{Mehrabi21}) have demonstrated the pernicious impact of low-data in face recognition \citep{Buolamwini18}, medical imaging \citep{Castro20}, and conservation \citep{Beery18} and spurious correlation in classification \citep{Geirhos19} and conservation \citep{Beery20}.
\cite{Arjovsky18} hypothesized that spurious correlation may be the underlying reason for poor generalization of models to unseen data.
To our knowledge, there has been no large scale work focused on understanding the benefits of different methods across these distribution shifts systematically across multiple datasets and with fine-grained control on the amount of shift.
Here we introduce a framework for creating these shifts in a controllable way to allow such challenges to be investigated robustly.

\paragraph{Benchmarking disentangled representations.} A related area, disentangled representation learning, aims to learn a representation where the factors of variation in the data are separated. If this could be achieved, then models should be able to generalise effortlessly to unseen data as investigated in multiple settings such as reinforcement learning \citep{Higgins17}. Despite many years of work in disentangled representations \citep{Higgins16,Burgess18,Kim18,Chen18b}, a benchmark study by \cite{Locatello19} found that, without supervision or implicit model or data assumptions, one cannot reliably perform disentanglement; however, weak supervision appears sufficient to do so \citep{Locatello20}.
\cite{Dittadi21,Schott21,Montero20} further investigated whether representations (disentangled or not) can interpolate, extrapolate, or compose properties; they found that when considering complex combinations of properties and multiple datasets, representations do not do so reliably.
% While \cite{Dittadi21} found that when considering one factor of variation, disentanglement could be used as a proxy for generalization, \cite{Schott21,Montero20} found that this was not the case when looking at more complex combinations of variations across a wider range of datasets and when the representation was not trained with the full data distribution.

\section{Conclusions}
This work has put forward a general, comprehensive framework to reason about distribution shifts.
We analyzed 19 different methods, spanning a range of techniques, over three distribution shifts -- {\em spurious correlation}, {\em low-data drift}, and {\em unseen data shift}, and two additional conditions -- {\em label noise} and {\em dataset size}.
We found that while results are not consistent across datasets and methods, a number of methods do better than an ERM baseline in some settings.
We then put forward a number of practical tips, promising directions, and open research questions.
We hope that our framework and comprehensive benchmark spurs research on in this area and provides a useful tool for practitioners to evaluate which methods work best under which conditions and shifts.

\subsubsection*{Acknowledgments}
The authors thank Irina Higgins and Timothy Mann for feedback and discussions while developing their work.
They also thank Irina, Rosemary Ke, and Dilan Gorur for reviewing earlier drafts.

\bibliography{longstrings,egbib}
\bibliographystyle{iclr2022_conference}

\appendix
\clearpage

\section{Overview}

In the appendix we give the full breakdown of results as well as additional results in \appref{sec:app:results}.
We then give a thorough literature review in \appref{sec:app:relatedwork}; a description of the datasets and how we set up the distribution shifts and conditions on these datasets in \appref{sec:app:dataset}; further details about each method evaluated in \appref{sec:app:methods}; and finally implementation details in \appref{sec:app:implementation}.

\section{Results}
\label{sec:app:results}
We give the complete breakdown of results for all methods and shifts in \appref{sec:app:results:complete}.
We give a detailed analysis on the impact of each augmentation type in RandAugment in \appref{sec:app:results:randaugment}.
Finally, we evaluate the performance using a different attribute as the label in \appref{sec:app:results:diffattributes} and using the ID (as opposed to OOD) validation set on \iwildcam and \camelyon in \appref{sec:app:results:idvsoood}.

\subsection{Complete results}
\label{sec:app:results:complete}
Here, we give the complete results over each dataset for each shift ({\em spurious correlation}, {\em low-data drift}, and {\em unseen data shift}) in \multfigref{fig:stronggeneralization}{fig:lowdatadrift}. In these plots, we plot the mean and standard deviation of each method on each dataset and sort them according to their performance.
The bars for each method are colored according to the general method they belong to (green denotes different architectures, orange different heuristic augmentation methods, and so on).

\subsection{A detailed analysis on the impact of augmentation}
\label{sec:app:results:randaugment}
We found that different methods of performing heuristic augmentation perform differently: some outperform the ERM baseline, some do not. 
Each of these methods is composed of individual augmentation techniques.
Here we investigate how each technique contributes to the end performance and thereby how the {\em choice} of augmentation function impacts the robustness of the models.

\begin{figure}[h]
    \centering
    \includegraphics[width=0.7\linewidth]{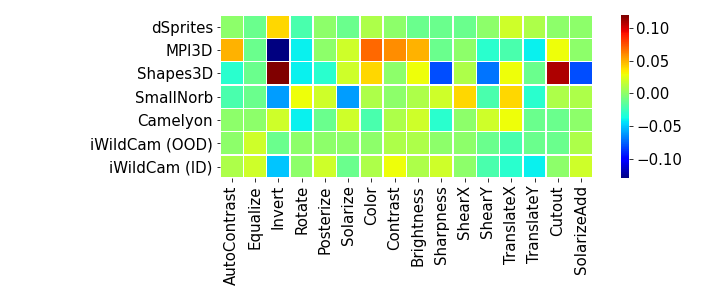}
    \caption{{\bf RandAugment ablation.} Performance of each augmentation across the different datasets. As can be seen, no one augmentation always improves performance. An augmentation that may improve performance on one dataset (e.g.~invert on \shapes) hurts performance on another (e.g.~invert on \smallnorb).}
    \label{fig:app:randaugment}
\end{figure}

We take the augmentations used in RandAugment. 
Instead of sampling all augmentations randomly, for each augmentation, we randomly sample the given augmentation or the identity function.
We plot the deviation from the mean in \figref{fig:app:randaugment} for each dataset under the setting {\em unseen data shift}.
The results are surprising.
No augmentation always leads to a strong boost in performance. 
For example, using {\em invert} improves performance on \dsprites, \shapes but harms performance on \mpithreed, \smallnorb, and \iwildcam.
Similarly, using {\em color} improves performance on most datasets but harms performance on \camelyon.

\subsection{Results with different attributes}
\label{sec:app:results:diffattributes}
\begin{figure}[h]
    \centering
    \includegraphics[width=0.25\linewidth]{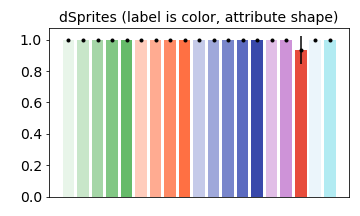}
    \caption{{\bf Shift 3: Unseen data shift.} We plot the top-1 accuracy (y-axis) for the \dsprites dataset. Unlike in the main paper, here the label is the color and the attribute the shape. Higher is better. All approaches solve the task.}
    \label{fig:app:dspritescolor}
\end{figure}

Here we investigate if the results are dependent on the attributes being investigated.
We investigate {\em unseen data shift} on \dsprites, but instead of predicting shape, we predict color.
We then leave out some shapes and test whether models can correctly predict color given the unseen shape.
We find that {\em all} methods generalise to the OOD case with approximately perfect score, as shown in \figref{fig:app:dspritescolor}.

\subsection{Results using different, OOD validation sets}
\label{sec:app:results:idvsoood}
\begin{figure}
    \centering
    \begin{subfigure}[b]{\linewidth}
    \includegraphics[width=\linewidth]{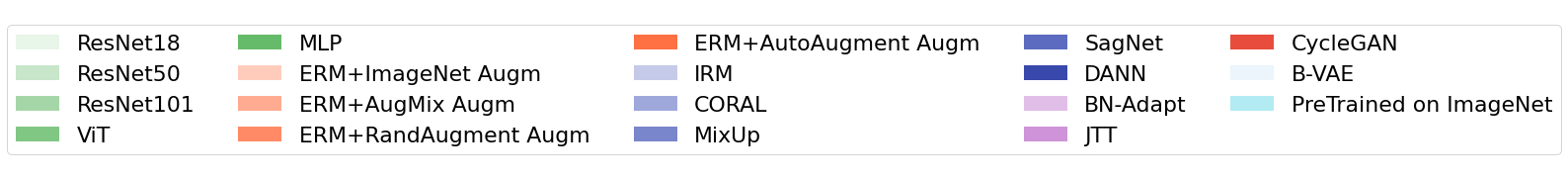}
    \end{subfigure} 
    
    \begin{subfigure}[b]{0.23\linewidth}
    \includegraphics[width=\linewidth]{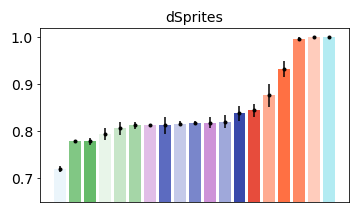}
    \caption{\dsprites}
    \end{subfigure} \hspace{0.1em}
    \begin{subfigure}[b]{0.23\linewidth}
    \includegraphics[width=\linewidth]{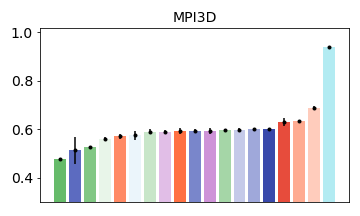}
    \caption{\mpithreed}
    \end{subfigure} \hspace{0.1em}
    \begin{subfigure}[b]{0.23\linewidth}
    \includegraphics[width=\linewidth]{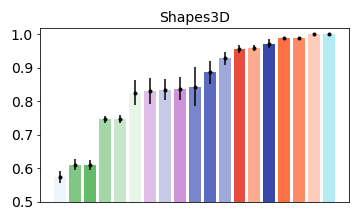}
    \caption{\shapes}
    \end{subfigure} \hspace{0.1em}
    \begin{subfigure}[b]{0.23\linewidth}
    \includegraphics[width=\linewidth]{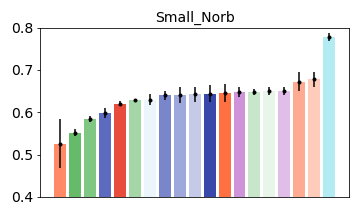}
    \caption{\smallnorb}
    \end{subfigure}
    
    \begin{subfigure}[b]{0.23\linewidth}
    \includegraphics[width=\linewidth]{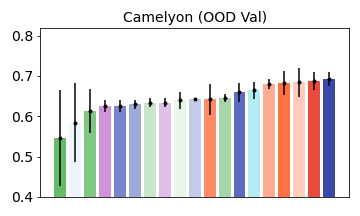}
    \caption{\camelyon. Using the ID val set.}
    \end{subfigure} \hspace{0.1em}
    \begin{subfigure}[b]{0.23\linewidth}
    \includegraphics[width=\linewidth]{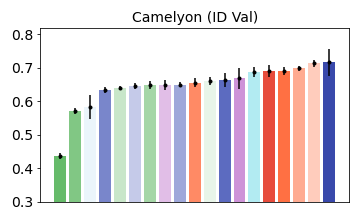}
    \caption{\camelyon. Using the OOD val set.}
    \end{subfigure} \hspace{0.1em}
    \begin{subfigure}[b]{0.23\linewidth}
    \includegraphics[width=\linewidth]{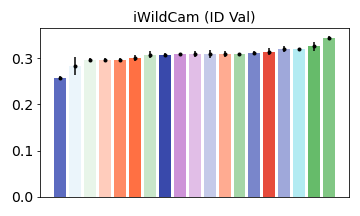}
    \caption{\iwildcam. Using the ID val set.}
    \end{subfigure}  \hspace{0.1em}
    \begin{subfigure}[b]{0.23\linewidth}
    \includegraphics[width=\linewidth]{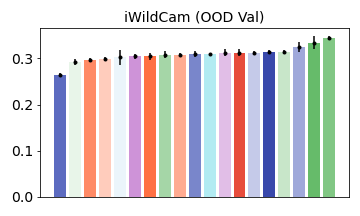}
    \caption{\iwildcam. Using the OOD val set.}
    \end{subfigure}
    \caption{{\bf Shift 3: Unseen data.} We plot the top-1 accuracy (y-axis) over different datasets. For each value of $N$ we resort the results and show them in order. Higher is better.}
    \label{fig:stronggeneralization}
\end{figure}

When evaluating on \iwildcam and \camelyon, we explore whether using the out-of-distribution (OOD) or in-distribution (ID) validation sets is preferable for obtaining best performance.
We find that neither the out-of-distribution (OOD) nor in-distribution (ID) validation sets perform best, but the relative performance remains similar.
The relative performance of different models in \figref{fig:stronggeneralization} when using the ID or OOD validation set is comparable on \camelyon and \iwildcam.
However, the maximum performance is not consistently better using either the ID or OOD set (for \camelyon, using the OOD set performs a bit better but for \iwildcam, using the ID set performs best).

\begin{figure}[ht]
    \centering
    \begin{subfigure}[b]{\linewidth}
    \includegraphics[width=\linewidth]{graphs/legend.png}
    \end{subfigure}
    \begin{subfigure}[b]{\linewidth}
    \includegraphics[width=\linewidth]{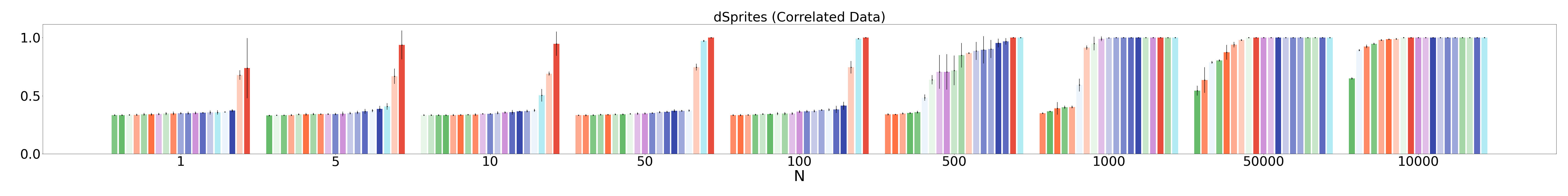}
    \caption{\dsprites}
    \end{subfigure}
    \begin{subfigure}[b]{\linewidth}
    \includegraphics[width=\linewidth]{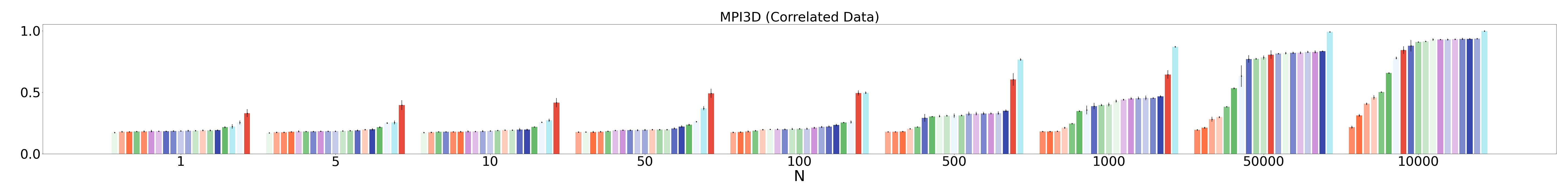}
    \caption{\mpithreed}
    \end{subfigure}
    \begin{subfigure}[b]{\linewidth}
    \includegraphics[width=\linewidth]{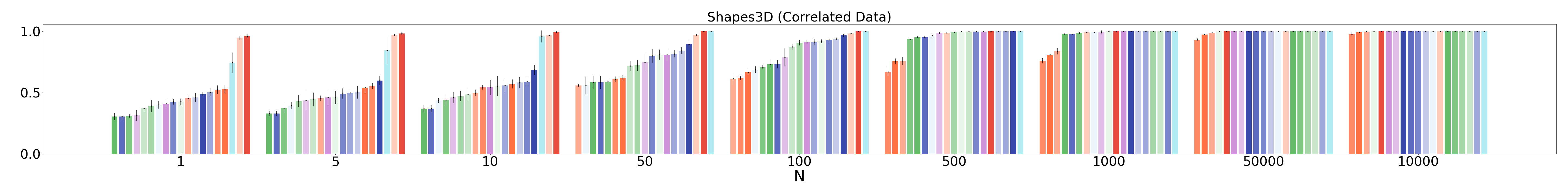}
    \caption{\shapes}
    \end{subfigure}
    \begin{subfigure}[b]{\linewidth}
    \includegraphics[width=\linewidth]{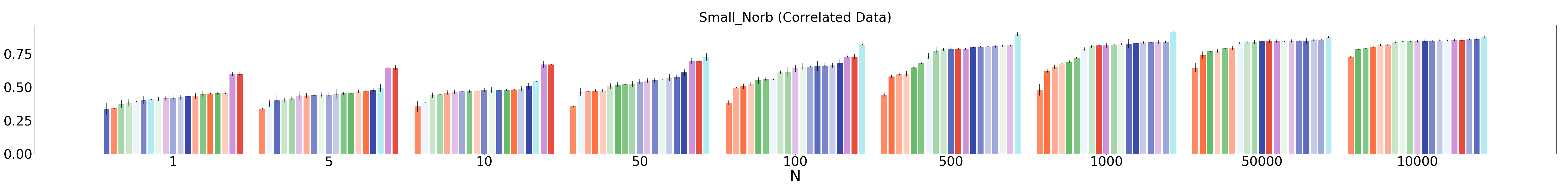}
    \caption{\smallnorb}
    \end{subfigure}
    \begin{subfigure}[b]{\linewidth}
    \includegraphics[width=\linewidth]{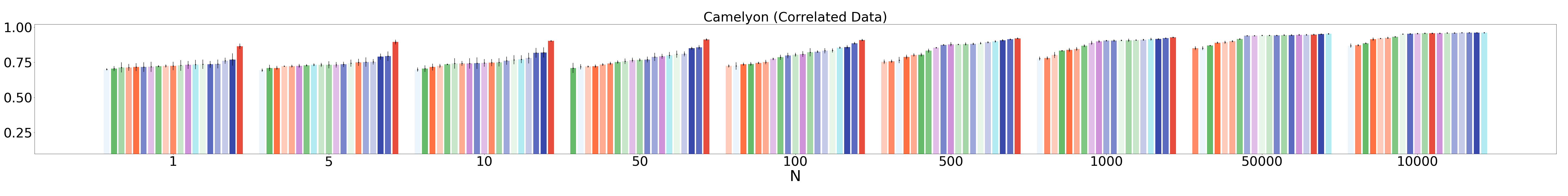}
    \caption{\camelyon}
    \end{subfigure}
    \begin{subfigure}[b]{\linewidth}
    \includegraphics[width=\linewidth]{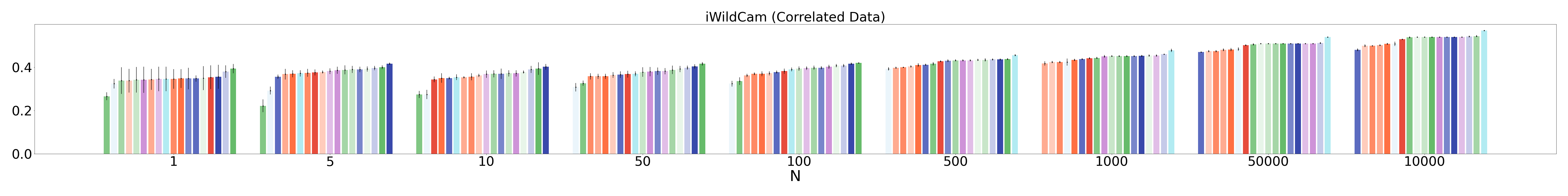}
    \caption{\iwildcam}
    \end{subfigure}
    \caption{{\bf Shift 1: Spurious correlation.} We plot the top-1 accuracy (y-axis) for different values of $N$ (the number of samples from the independent distribution, the x-axis) over different datasets. For each value of $N$ we resort the results and show them in order. Higher is better.}
    \label{fig:spuriouscorrelation}
\end{figure}

\begin{figure}[ht]
    \centering
    \begin{subfigure}[b]{\linewidth}
    \includegraphics[width=\linewidth]{graphs/legend.png}
    \end{subfigure}
    \begin{subfigure}[b]{\linewidth}
    \includegraphics[width=\linewidth]{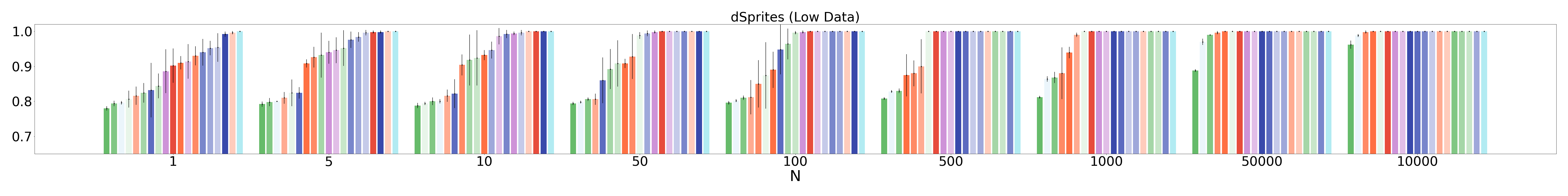}
    \caption{\dsprites}
    \end{subfigure}
    \begin{subfigure}[b]{\linewidth}
    \includegraphics[width=\linewidth]{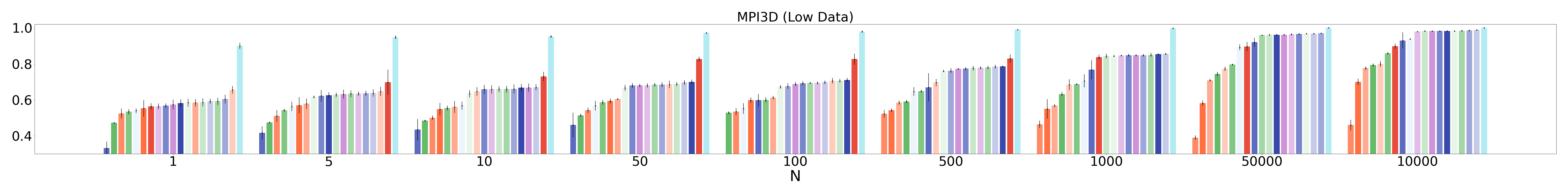}
    \caption{\mpithreed}
    \end{subfigure}
    \begin{subfigure}[b]{\linewidth}
    \includegraphics[width=\linewidth]{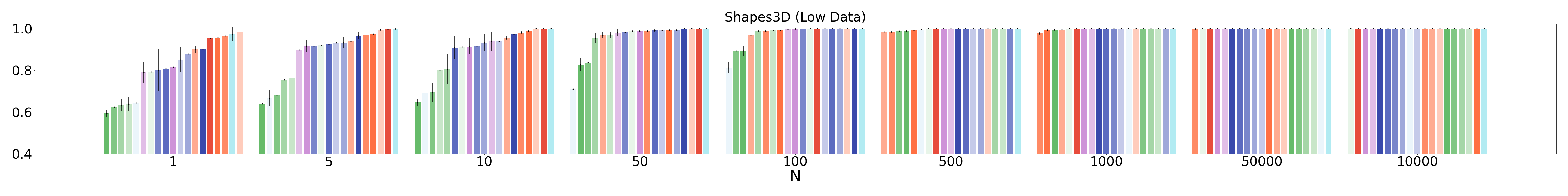}
    \caption{\shapes}
    \end{subfigure}
    \begin{subfigure}[b]{\linewidth}
    \includegraphics[width=\linewidth]{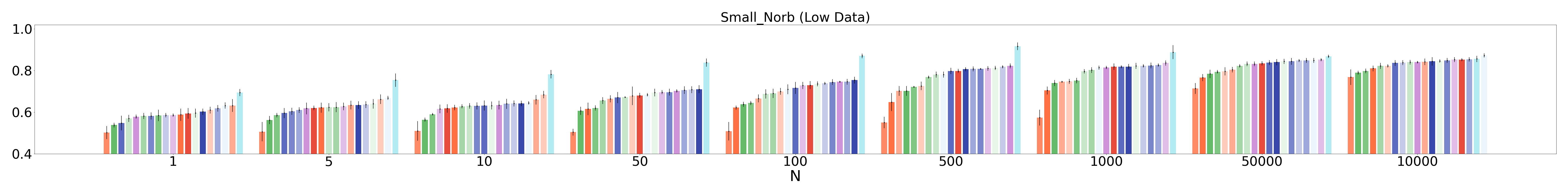}
    \caption{\smallnorb}
    \end{subfigure}
    \begin{subfigure}[b]{\linewidth}
    \includegraphics[width=\linewidth]{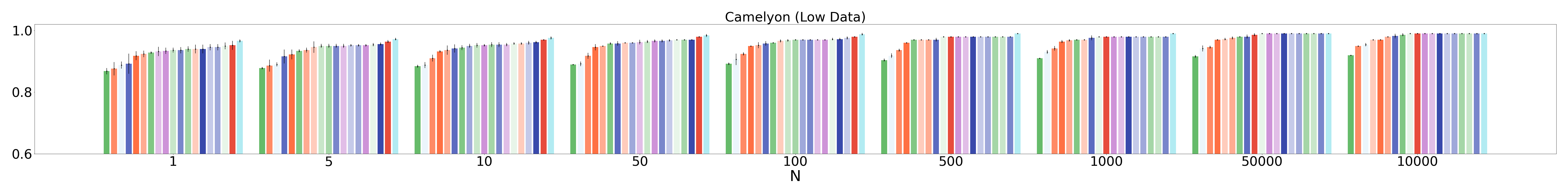}
    \caption{\camelyon}
    \end{subfigure}
    \begin{subfigure}[b]{\linewidth}
    \includegraphics[width=\linewidth]{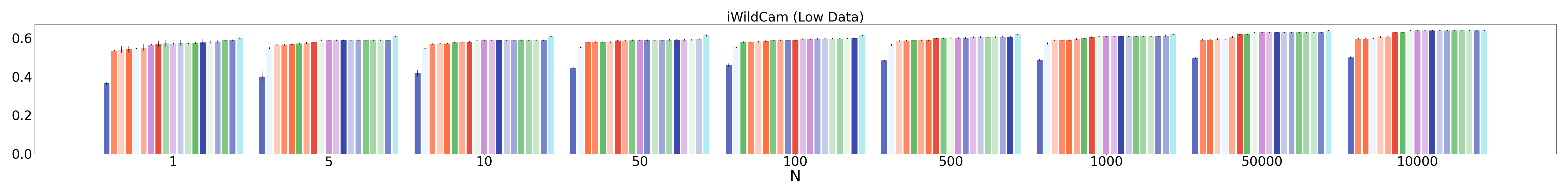}
    \caption{\iwildcam}
    \end{subfigure}
    \caption{{\bf Shift 2: Low data drift.}  We plot the top-1 accuracy (y-axis) for different values of $N$ (the number of samples from the independent distribution, the x-axis) over different datasets. For each value of $N$ we resort the results and show them in order. Higher is better.}
    \label{fig:lowdatadrift}
\end{figure}

\clearpage

\section{Literature review}
\label{sec:app:relatedwork}
Here we discuss in depth related literature.
We first discuss datasets used to evaluate different distribution shifts.
We then discuss methods related to the five common approaches we consider in the paper: architecture choice, data augmentation, domain generalization methods, adaptive approaches, and representation learning.
We additionally discuss the hypotheses present in the current literature and whether our results corroborate or dispute those hypotheses.

\paragraph{Datasets to evaluate distribution shifts.}
Obtaining datasets of real world distribution shifts is challenging and expensive.
As a result, many works have focussed on using synthetic or existing datasets to build their benchmarks.
One approach is to create synthetic shifts over ImageNet to study how models generalise in specific, synthetic conditions: ImageNet-C \citep{Hendrycks19b} studies the impact of standard corruptions; Stylized ImageNet \citep{Geirhos19} studies the impact of texture; Waterbirds and the Backgrounds challenge \citep{Sagawa20,Xiao20} study the impact of a fake background; and colored MNIST \citep{Gulrajani20} studies the impact of color on classification.
Another approach is to consider how models trained on a given dataset generalise to other datasets of the same set of objects (e.g.~cartoons to paintings in PACS or ImageNet to ImageNetV2) \citep{Torralba11,Li17,Recht19,Peng19,Venkateswara17} or over time \citep{Shankar19}.
While these datasets give insight into the biases of a trained model, models that do well on them may not necessarily do well on real world tasks.
As a result, the WILDS dataset \citep{Wilds2020} was created as a collection of real world distribution shifts where the OOD task is well defined (e.g.~the model should generalise to unseen countries for satellite imagery or hospitals for tumour identification) in order to evaluate progress.

Our framework is complementary to these datasets (\tabref{tab:benchmarks}).
Given a dataset, our framework can be used to set up a desired distribution shift to be investigated with fine-grained control over the type of distribution shift and amount of shift.
Moreover, our framework provides extendable classes for a wide range of approaches and datasets.
Finally, we present a robustness framework inspired by disentangling literature analyzing {\em when} we expect models to be able to generalise across these shifts.

\paragraph{Frameworks to evaluate the impact of label noise.}
A related area of work evaluates the impact of label noise on downstream performance. 
To ablate the impact of label noise, this was originally studied by constructing synthetic noise on standard datasets \citep{Han18,Hendrycks18,Khetan18,Patrini17} in a variety of conditions: some labels are trusted \citep{Hendrycks18}, there is a fixed label budget \citep{Khetan18}, or there is knowledge of the confusion matrix \citep{Patrini17}.
\cite{Gu21} generalised this framework to account for auxiliary information (such as rater expertise) and the varying difficulty of samples.
Our framework mostly focuses on distribution shifts but can be extended to include more complex types of label noise.

\begin{table}
\begin{tabular}{c|c|c|c} \toprule
     & Controllable shifts? & Many methods? & Real world motivated? \\ \midrule
     Ours & \cmark & \cmark & \cmark \\
     \wilds \citep{Wilds2020} & \xmark & \xmark & \cmark \\
     \cite{Gulrajani20} & \xmark & \cmark & \xmark \\
     \cite{Hendrycks21} & \xmark & \cmark & \xmark \\ \bottomrule

\end{tabular}

     \caption{Comparison of the three most similar benchmark works to ours. Unlike these works, our framework allows for controlling the  distribution shift to be analysed. Moreover, we use real-world motivated datasets and evaluate methods across multiple approaches; we encompass more  types of approaches than \cite{Gulrajani20} and \cite{Hendrycks21}.}
     \label{tab:benchmarks}
\end{table}

\paragraph{Architecture choice.}
Previous work has investigated the impact of model size on robustness to various distribution shifts, finding that larger models generally are more robust when considering common corruptions \citep{Hendrycks19} and adversarial training \citep{Xie20}.
However, \citep{Schott21} finds that larger models do not give representations that generalise better within the same domain.
We find that there is no strict rule correlating model size or model capacity with performance. Sometimes deeper ResNets perform best, sometimes not.
The ViT model (with the highest capacity) often performs worse, but on \iwildcam it performs best.
However, we note that using additional data or pretraining on all models or methods may change their relative performance.

\paragraph{Data augmentation.}
Data augmentation is often pivotal to achieve state of the art performance on machine learning benchmarks, leading to a wide range of methods to perform heuristic data augmentation \citep{He16,Hendrycks20b,Cubuk19,Cubuk20,Zhang18}.
When considering domain generalization,  more specific techniques for augmentation have been devised to improve a model's ability to generalise using MixUp \citep{Gulrajani20,Xu20,Yan20,Wang20} or knowledge of the domains \citep{Zhang19b}.

Instead of using heuristic data augmentations which cannot be used to learn more complex transformations, the desired augmentation can be learnt.
The training data can be augmented by transforming samples using a generative model to either come from another part of the domain conditioned on an attribute \citep{Goel20}, another domain \citep{Zhou20}, be domain agnostic \citep{Maria19}, or to have a different style \citep{Gowal20,Geirhos19}.
These methods often build on work in image generation, such as \cyclegan \citep{Zhu17} and \stylegan \citep{Karras19}.

We find that these approaches can be effective to learn invariance to the heuristic or learned transformation.
However, performance depends on whether the augmentation is a useful property to learn invariance against, else the model may waste capacity.
When learning the augmentation, performance is limited by the quality of the learned generative model.

\paragraph{Domain generalization.}
While many works can be viewed as aiming to improve robustness of models in new domains, here we focus on those methods that aim to learn domain invariant features in stochastically  trained machine learning models.
One impactful approach to learning features invariant to the domain is DANN \citep{Ganin16} (domain adversarial neural networks). 
This work uses an adversarial network \citep{Goodfellow14} to enforce that the features cannot be predictive of the domain.
Later work considered a number of ways to enforce invariance, such as the following: minimizing the maximum mean discrepancy \citep{Long15,Long17}, invariance of the conditional distribution \citep{Li18}, and invariance of the covariance matrix of the feature distribution \citep{Sun16}.
However, enforcing invariance is challenging and often too strict \citep{Johansson19,Zhao19}.
As a result, IRM \citep{Arjovsky18} instead enforces that the optimal classifier for different domains is the same.
We find that DANN performs consistently best but that performance varies over different datasets and distribution shifts.

\paragraph{Adaptive approaches.}
Instead of learning a single model and treating samples similarly, adaptive approaches can be used to either modify model parameters or dynamically modify training if there are multiple domains (or groups with different attributes) within the training set.
Ways to adapt the model parameters to a new domain include meta learning (as in MAML \citep{Finn17}) and adapting the batch normalisation statistics based on the different domains (as in BN-Adapt \citep{Schneider20}).
Instead of adapting the parameters, other approaches reweigh the importance of samples on which the model struggles.
This, intuitively, should force the model to spend more capacity on more challenging parts of the domain (or groups with fewer samples).
In GroupDRO \citep{Sagawa20}, this amounts to putting more mass on samples from the more challenging domains at train time using the loss to determine the challenging domains.
In JTT \citep{Liu21}, this is done by two stage training. First, a classifier is trained in a standard manner.
Then, a second classifier is trained by upweighting the samples with a high loss according to the first classifier. The authors posit that the most challenging samples will be those coming from groups with few samples (the low-data regions).
We find that neither JTT nor BN-Adapt consistently perform better than the baselines.

\paragraph{Representation learning.}
Finally, instead of focusing on the downstream task, another approach is to learn a representation $p(z|\mI)$ that approximates the true prior over the latent factors.
This can be done by pretraining on large amounts of data such as ImageNet \citep{Russakovsky15,Deng09}, as explored by \citep{Hendrycks21} for domain generalization, such that the learned representation is potentially more robust and general.
Our results corroborate their findings: in many cases pretraining is helpful.
However, this is not universally true.
On \camelyon and \iwildcam, pretraining did not improve performance across all shifts.
For example, pretraining was unhelpful under spurious correlation on \camelyon.

Another approach is to learn a representation subject to constraints.
This is what the disentanglement literature aims to do: find a representation that is sparse and low dimensional, which hopefully will correspond to a higher level, disentangled representation.
Disentanglement is a large research area, so we briefly mention some formative works, such as the $\beta$-VAE \citep{Higgins16,Burgess18} and FactorVAE \citep{Kim18}, which build on the original VAE \citep{Kingma13,Rezende14} formulation. 
Other approaches build on GANs \citep{Goodfellow14}, such as InfoGAN \citep{Chen16}.
A recent study of these methods \citep{Locatello19} found that they did not reliably disentangle the representation into a semantically meaningful set of latent variables.
Therefore, it is unclear how robust these methods are when used for distribution shifts, but our results imply that more research is needed to make these approaches practically useful for robustness.

\section{Dataset}
\label{sec:app:dataset}
Here we give further details about the datasets in \appref{sec:app:datasetdetails}, how we set up the shifts and conditions for these datasets in \appref{sec:app:datasetshifts} and finally we give samples from the different datasets and shifts in \appref{sec:app:datasamples}.

\subsection{Details}
\label{sec:app:datasetdetails}
Here we give further detail about each of the datasets and how we set the two attributes: $y^l, y^a$.

\paragraph{\dsprites.} \dsprites \citep{Matthey17} consists of shapes (heart, ellipse, and square), which we augment with three colors (red, green, and blue). The shapes vary in location, scale, and orientation.
We make the shape the label, and the attribute the color (we consider other choices in \appref{sec:app:results:diffattributes}).

\paragraph{\mpithreed.} \mpithreed \citep{Gondal19} consists of real images of shapes on a robotic arm. There are six shapes and the images vary in terms of the object color, object size, camera height, background color and the rotation about the horizontal and vertical axis.
We make the shape the label and object color the other attribute.

\paragraph{\shapes.} \shapes \citep{Burgess18b} consists of images of shapes centered in a synthetic room. There are three shapes and the images vary in terms of the floor hue, object hue, orientation, scale, shape, and wall hue.
We make the shape the label and object color the other attribute.

\paragraph{\smallnorb.} \smallnorb \citep{LeCun04} consists of images of toys of five categories with varying lighting, elevation and azimuths.
The aim is to create methods that generalize to unseen samples of the five categories.
We make the object category the label and azimuth the other attribute for {\em low-data drift} and {\em unseen data shift}. We make the lighting the other attribute for {\em spurious correlation} and under the noise and fixed data size conditions.

\paragraph{\camelyon.} \camelyon \citep{Wilds2020,Bandi18} contains tumour cell images coming from different hospitals.
We follow \cite{Wilds2020} and make the label the presence of tumor cells and the other attribute the hospital from which the image came from.
In the {\em spurious correlation, low-data drift} settings, we resplit the dataset randomly (there are no held out hospitals).
In the {\em spurious correlation} setting, we correlate the presence of tumor cells with the hospital (all images from a given hospital either do or do not have tumors).
In the {\em low-data} setting, we select the hospitals that are out of distribution for \cite{Wilds2020} as the hospitals for which we only have $N$ samples.
In the {\em unseen data shift}, we use the split given by \cite{Wilds2020} and optionally use either the in-distribution or out-of-distribution validation set for model selection.

\paragraph{\iwildcam.}
\iwildcam \citep{Wilds2020,Beery18} contains camera trap imagery.
There are a set of locations. The camera is kept in the same fixed spot in each location, and takes photos at different times.
The task is to determine if there is an animal in the image and which animal is present.
In the {\em spurious correlation, low-data drift} settings, we resplit the dataset randomly (there are no held out locations).
In the {\em spurious correlation} setting, we correlate the presence of a given animal with a location (all images from a given location {\em only} show a given animal).
In the {\em low-data drift} setting, we select the locations that are out of distribution for \cite{Wilds2020} as the locations for which we only have $N$ samples.
In the {\em unseen data shift} setting, we use the split given by \cite{Wilds2020} and optionally use either the in-distribution or out-of-distribution validation set for model selection.

\subsection{Evaluated shifts and conditions}
\label{sec:app:datasetshifts}
We further describe how we set up the shifts in this section and define each of the shifts for each dataset in \tabref{tab:app:datashifts}.
Note that the amount of correlation, total number of samples from the low-data region, and the probability of sampling from the low-data or correlated distributions are controllable in our framework. 
Additionally, we can control the amount of label noise and total dataset size.

\label{sec:expshift1}
\paragraph{Shift 1: Spurious correlation.} Under spurious correlation, we correlate $y^l, y^a$.
At test time, these attributes are uncorrelated.
We vary the amount of correlation by creating a new dataset with all samples from the correlated distribution in the dataset and $N$ samples from the uncorrelated distribution; this forms the training set.
We set $N >= 1$ (as if $N = 0$, then the problem is ill defined as to what is the correct label).
The test set is composed of sampling from the uncorrelated distribution and is disjoint from the training samples.

\paragraph{Shift 2: Low-data drift.} 
\label{sec:expshift2}
Under low-data drift, 
we consider the set $\attributeSet{a}$.
For some subset $\attributeSet{a}_c \subset \attributeSet{a}$, we only see $N$ samples of those attributes. 
For all other values of $\attribute{a}$ ($\attributeSet{a} / \attributeSet{a}_c$), the model has access to all samples.

\paragraph{Shift 3: Unseen data shift}
\label{sec:expshift3}
This is a special case of {\em low-data drift}, where we set $N = 0$.

\paragraph{Condition 1: Noisy labels.} 
To investigate how methods perform in the presence of noise, we add uniform label noise with increasing probability.
We take the low-data setting and fix the value of $N$.
We then vary the amount of noise $p$.

\paragraph{Condition 2: Dataset size.} We investigate how performance degrades as the total size of the training dataset changes.
We again take the low-data setting  but we vary the total number $n$ of samples from the train set and fix the ratio $\frac{N}{n}$.

\begin{table}[b]
    \scriptsize
    \centering
    \begin{tabular}{cp{1.6cm}p{1.8cm}|p{2.3cm}p{2.3cm}p{0.8cm}p{0.8cm}}
         Dataset & label & nuisance attr & SC & LD / UDS \newline ($\attributeSet{a}_c$) & Noise \newline ($N$) & Dataset size \newline ($N/n$) \\ \toprule
         \dsprites & $l$ = shape \newline $v_i \in \attributeSet{l}$ \newline $|\attributeSet{l}| = 3$ & $a$ = color \newline $u_i \in \attributeSet{a}$ \newline $|\attributeSet{a}| = 3$ & $v_i \sim u_i$  & $\{\text{blue}\}$ &  $10$ & $0.001$\\ \midrule
         \mpithreed & $l$ = shape \newline $v_i \in \attributeSet{l}$ \newline $|\attributeSet{l}| = 6$ & $a$ = color \newline $u_i \in \attributeSet{a}$ \newline $|\attributeSet{a}| = 6$ & $v_i \sim u_i$  & $\{u_i| i > 2\}$ & $10$ & $ 0.001$\\ \midrule
         \shapes & $l$ = shape \newline $v_i \in \attributeSet{l}$ \newline $|\attributeSet{l}| = 4$ & $a$ = obj.~color \newline $u_i \in \attributeSet{a}$ \newline $|\attributeSet{a}| = 10$ & $v_i \sim u_i, i \leq 4$  & $\{u_i| i > 2\}$ & $10$ & $0.001$\\ \midrule
         \smallnorb & $l$ = category \newline $v_i \in \attributeSet{l}$ \newline $|\attributeSet{l}| = 5$ & $a$ = azimuth \newline $u_i \in \attributeSet{a}$ \newline $|\attributeSet{a}| = 18$ \newline \newline $b$ = lighting \newline $w_i \in \attributeSet{b}$ \newline $|\attributeSet{b}| = 6$ & $v_i \sim w_i, i \leq 5$  & $\{u_i| i > 4\}$ & $10$ & $0.001$\\ \midrule
         \camelyon & $l$ = tumor \newline $v_i \in \attributeSet{l}$ \newline $|\attributeSet{l}| = 2$ & $a$ = hospital \newline $u_i \in \attributeSet{a}$ \newline $|\attributeSet{a}| = 5$ & $v_i=0 \sim u_{i \in \text{[1,2,3]} }$ \newline $v_i=1 \sim u_{i \in \text{[4,5]}}$  & $\{u_i| i \in \{2, 3\} \}$  & $10$ & $0.001$\\ \midrule
         \iwildcam & $l$ = animal \newline $v_i \in \attributeSet{l}$ \newline $|\attributeSet{l}| = 186$ & $a$ = location \newline $u_i \in \attributeSet{a}$ \newline $|\attributeSet{a}| = 324$ & $v \sim u$  & $|\attributeSet{a}_c| = 79$ & $10$ & $0.001$\\ \midrule
    \end{tabular}
    \caption{The precise dataset shifts we use for each dataset. We describe the label and nuisance attribute for each dataset. We additionally describe the setup for {\em spurious correlation (SC)} setting and for the {\em low-data drift (LD)/unseen data shift (UDS)} settings. Finally, we give the hyperparameters set in the {\em LD} setting when evaluating under the two additional conditions: label noise and fixed dataset size. $\sim$ denotes correlation. We denote each value in the attribute sets using 1-based indexing. For \iwildcam, not all locations have all animals, so we find the most commonly occurring location for each animal and correlate that animal with that location. Additionally, for \iwildcam and \camelyon, we use the OOD set from \cite{Wilds2020} as $\attributeSet{a}_c$. For \iwildcam, there are $72$ locations.}
    \label{tab:app:datashifts}
\end{table}

\subsection{Samples from the different distributions}
\label{sec:app:datasamples}
\begin{figure}
    \centering
    \begin{subfigure}[b]{0.48\linewidth}
    \includegraphics[width=\linewidth]{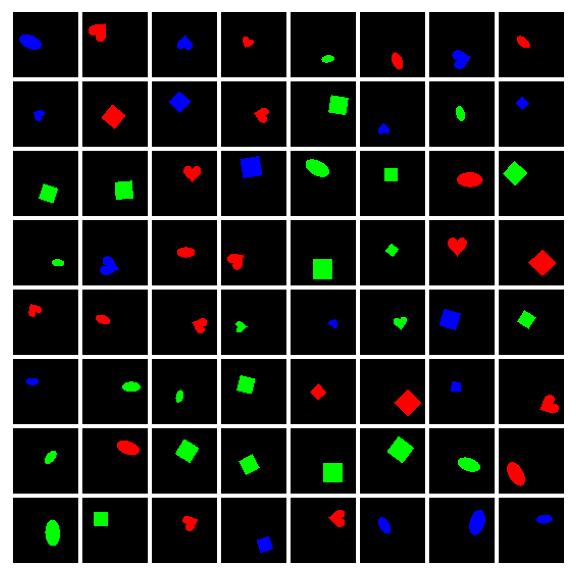}
    \caption*{\dsprites (IID test).}
    \end{subfigure}
    \begin{subfigure}[b]{0.48\linewidth}
    \includegraphics[width=\linewidth]{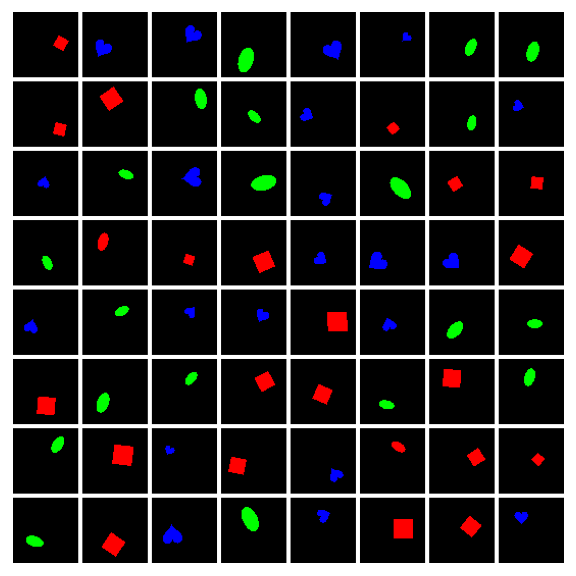}
    \caption*{\dsprites ({\bf Shift 1: Spurious correlation.})} 
    \end{subfigure}
    
    \begin{subfigure}[b]{0.48\linewidth}
    \includegraphics[width=\linewidth]{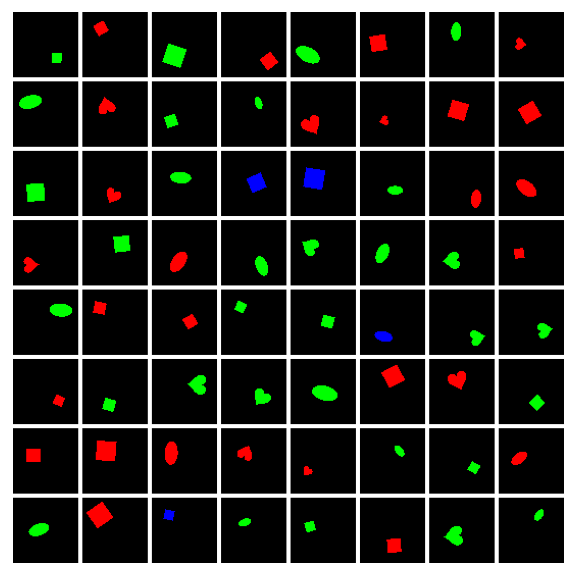}
    \caption*{\dsprites ({\bf Shift 2: Low-data drift.})} 
    \end{subfigure}
    \begin{subfigure}[b]{0.48\linewidth}
    \includegraphics[width=\linewidth]{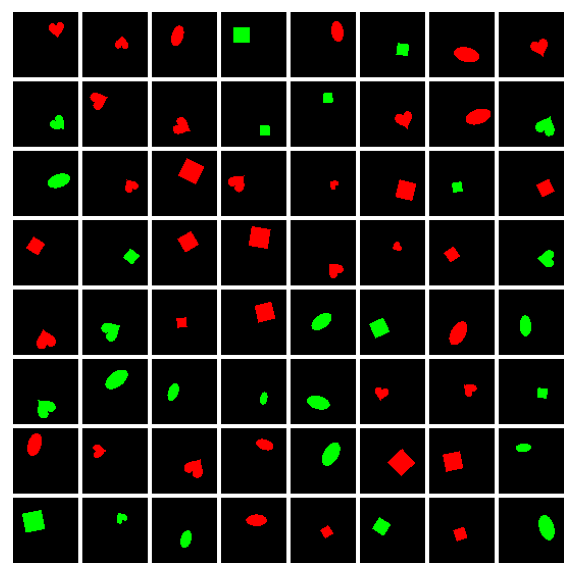}
    \caption*{\dsprites ({\bf Shift 3: Unseen data shift.})} 
    \end{subfigure}
    \caption{Sample distributions on \dsprites.}
    \label{fig:dataset:dsprites}
\end{figure}

\begin{figure}
    \centering
    \begin{subfigure}[b]{0.48\linewidth}
    \includegraphics[width=\linewidth]{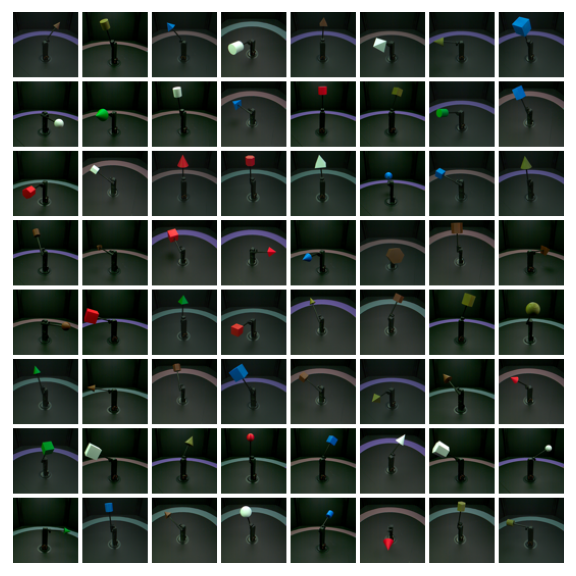}
    \caption*{\mpithreed (IID test).}
    \end{subfigure}
    \begin{subfigure}[b]{0.48\linewidth}
    \includegraphics[width=\linewidth]{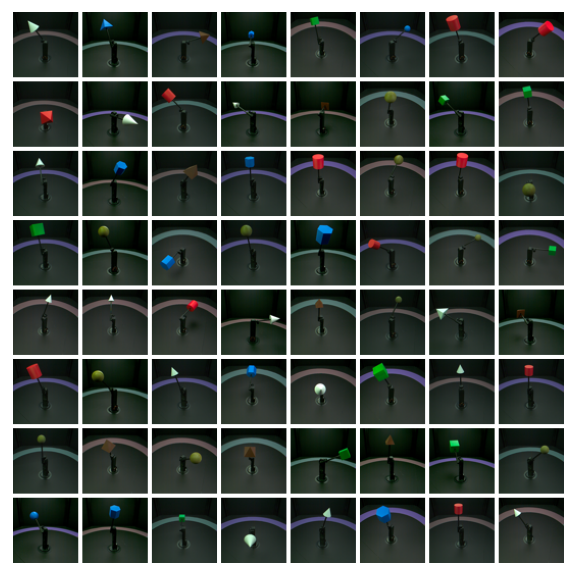}
    \caption*{\mpithreed ({\bf Shift 1: Spurious correlation.})} 
    \end{subfigure}
    
    \begin{subfigure}[b]{0.48\linewidth}
    \includegraphics[width=\linewidth]{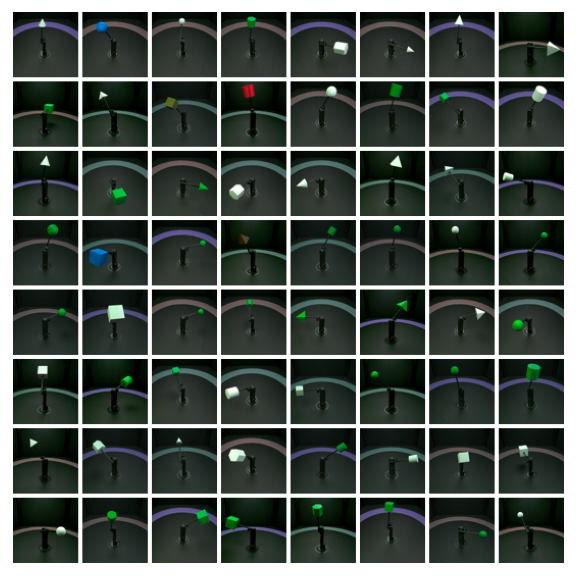}
    \caption*{\mpithreed ({\bf Shift 2: Low-data drift.})} 
    \end{subfigure}
    \begin{subfigure}[b]{0.48\linewidth}
    \includegraphics[width=\linewidth]{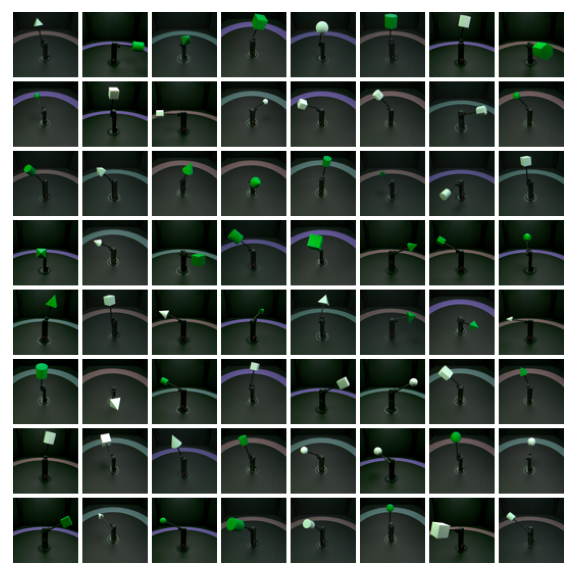}
    \caption*{\mpithreed ({\bf Shift 3: Unseen data shift.})} 
    \end{subfigure}
    \caption{Sample distributions on \mpithreed.}
    \label{fig:dataset:mpi3d}
\end{figure}

\begin{figure}
    \centering
    \begin{subfigure}[b]{0.48\linewidth}
    \includegraphics[width=\linewidth]{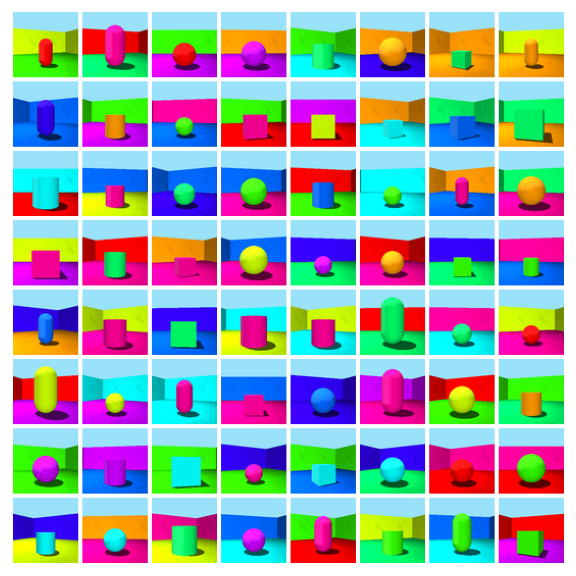}
    \caption*{\shapes (IID test).}
    \end{subfigure}
    \begin{subfigure}[b]{0.48\linewidth}
    \includegraphics[width=\linewidth]{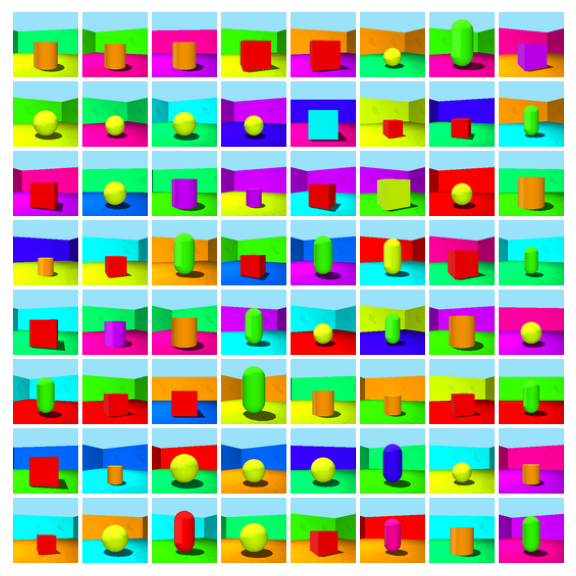}
    \caption*{\shapes ({\bf Shift 1: Spurious correlation.})} 
    \end{subfigure}
    
    \begin{subfigure}[b]{0.48\linewidth}
    \includegraphics[width=\linewidth]{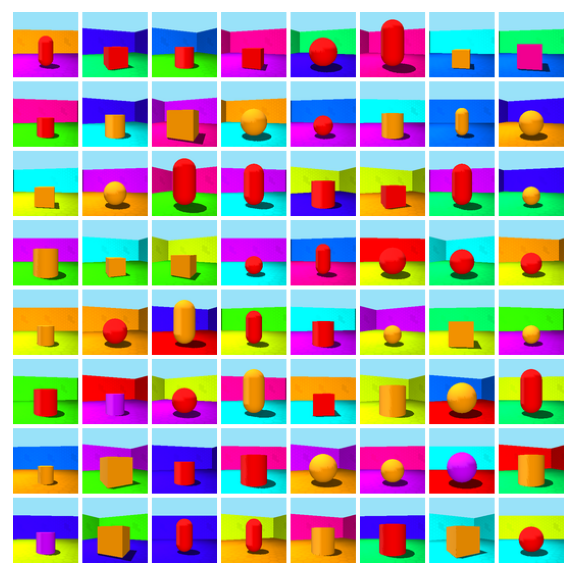}
    \caption*{\shapes ({\bf Shift 2: Low-data drift.})} 
    \end{subfigure}
    \begin{subfigure}[b]{0.48\linewidth}
    \includegraphics[width=\linewidth]{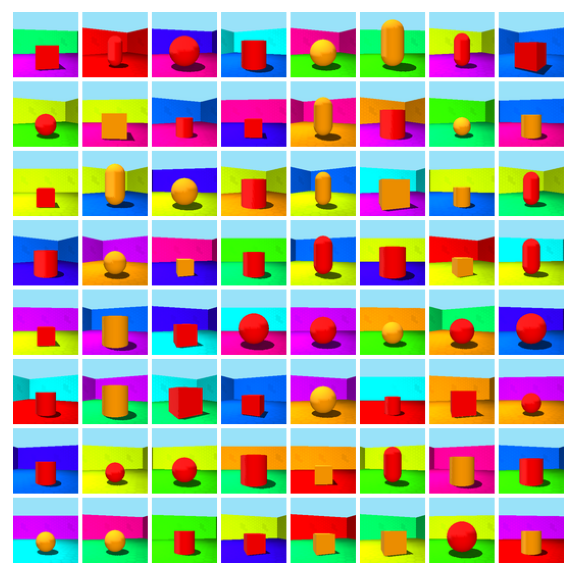}
    \caption*{\shapes ({\bf Shift 3: Unseen data shift.})} 
    \end{subfigure}
    \caption{Sample distributions on \shapes.}
    \label{fig:dataset:shapes}
\end{figure}

\begin{figure}
    \centering
    \begin{subfigure}[b]{0.48\linewidth}
    \includegraphics[width=\linewidth]{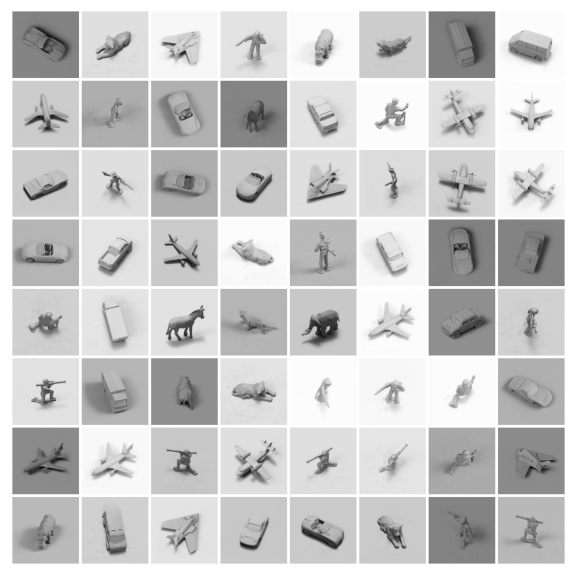}
    \caption*{\smallnorb (IID test).}
    \end{subfigure}
    \begin{subfigure}[b]{0.48\linewidth}
    \includegraphics[width=\linewidth]{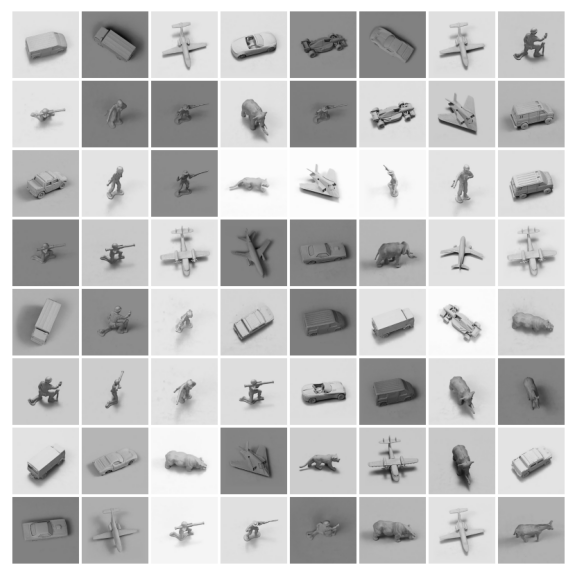}
    \caption*{\smallnorb ({\bf Shift 1: Spurious correlation.})} 
    \end{subfigure}
    
    \begin{subfigure}[b]{0.48\linewidth}
    \includegraphics[width=\linewidth]{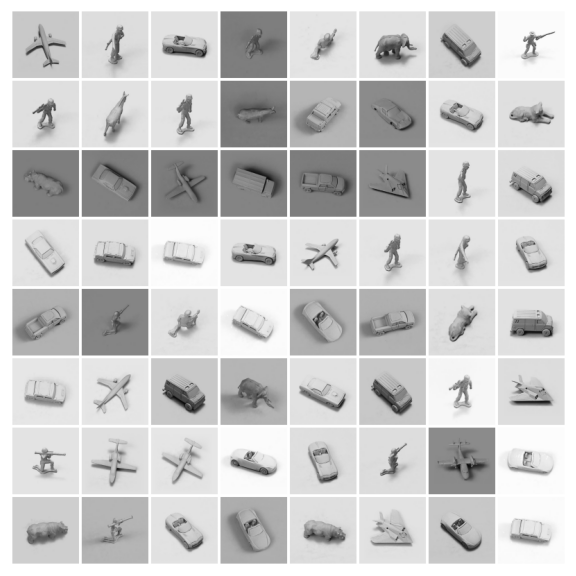}
    \caption*{\smallnorb ({\bf Shift 2: Low-data drift.})} 
    \end{subfigure}
    \begin{subfigure}[b]{0.48\linewidth}
    \includegraphics[width=\linewidth]{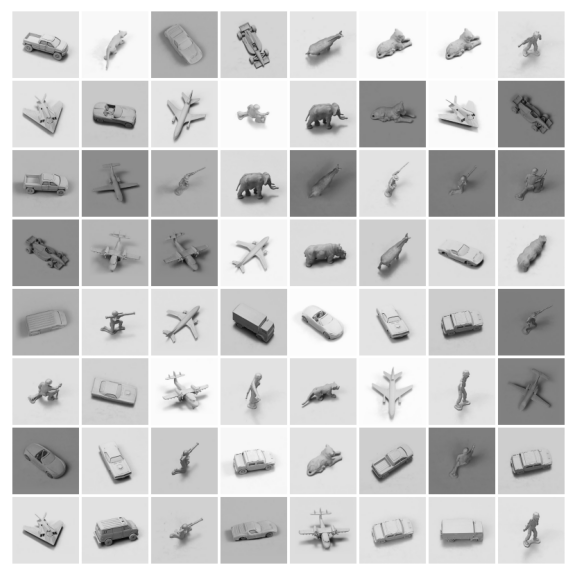}
    \caption*{\smallnorb ({\bf Shift 3: Unseen data shift.})} 
    \end{subfigure}
    \caption{Sample distributions on \smallnorb.}
    \label{fig:dataset:smallnorb}
\end{figure}

\begin{figure}
    \centering
    \begin{subfigure}[b]{0.48\linewidth}
    \includegraphics[width=\linewidth]{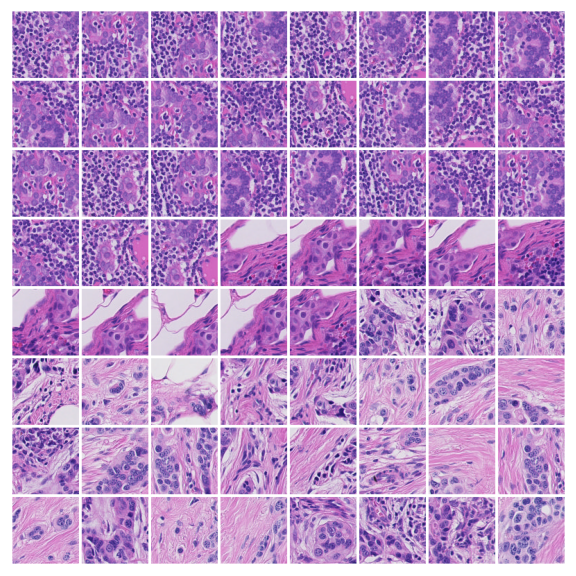}
    \caption*{\camelyon (OOD test).}
    \end{subfigure}
    \begin{subfigure}[b]{0.48\linewidth}
    \includegraphics[width=\linewidth]{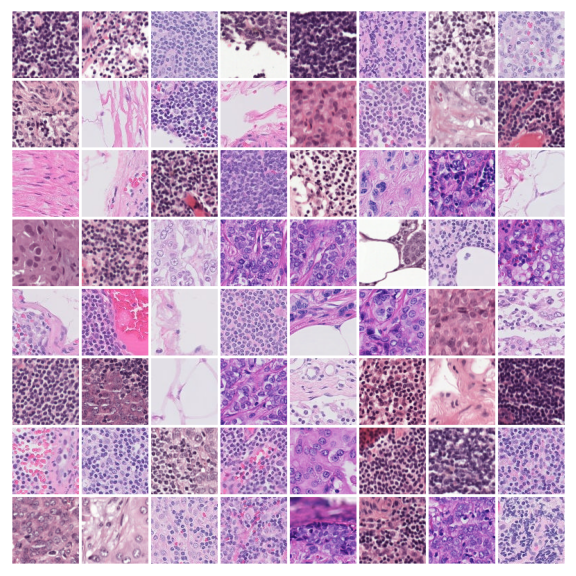}
    \caption*{\camelyon ({\bf Shift 1: Spurious correlation.})} 
    \end{subfigure}
    
    \begin{subfigure}[b]{0.48\linewidth}
    \includegraphics[width=\linewidth]{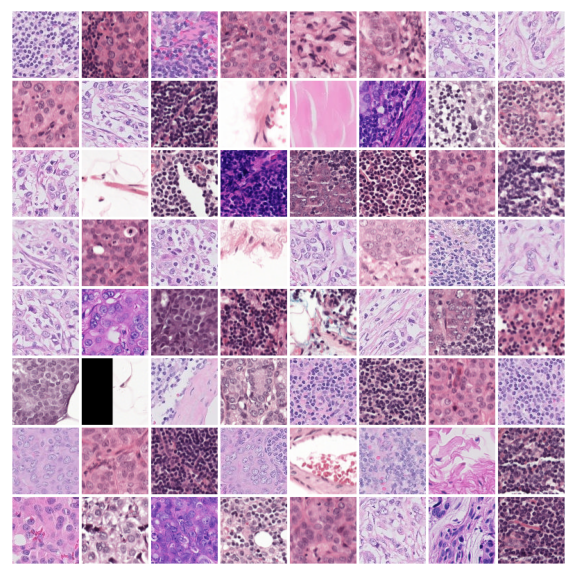}
    \caption*{\camelyon ({\bf Shift 2: Low-data drift.})} 
    \end{subfigure}
    \begin{subfigure}[b]{0.48\linewidth}
    \includegraphics[width=\linewidth]{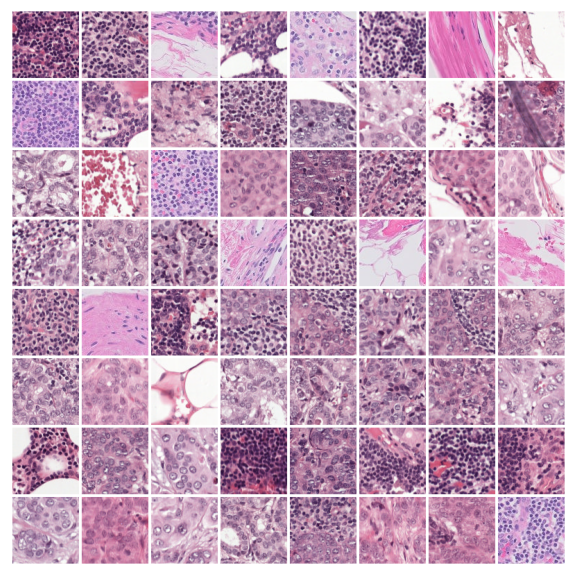}
    \caption*{\camelyon ({\bf Shift 3: Unseen data shift.})} 
    \end{subfigure}
    \caption{Sample distributions on \camelyon.}
    \label{fig:dataset:camelyon}
\end{figure}

\begin{figure}
    \centering
    \begin{subfigure}[b]{0.48\linewidth}
    \includegraphics[width=\linewidth]{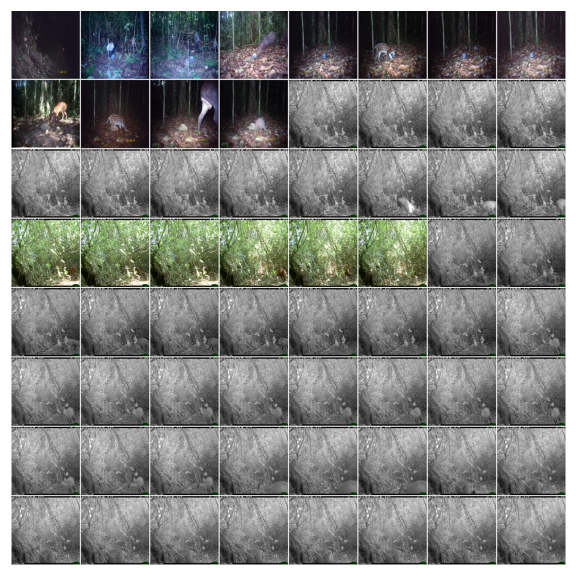}
    \caption*{\iwildcam (OOD test).}
    \end{subfigure}
    \begin{subfigure}[b]{0.48\linewidth}
    \includegraphics[width=\linewidth]{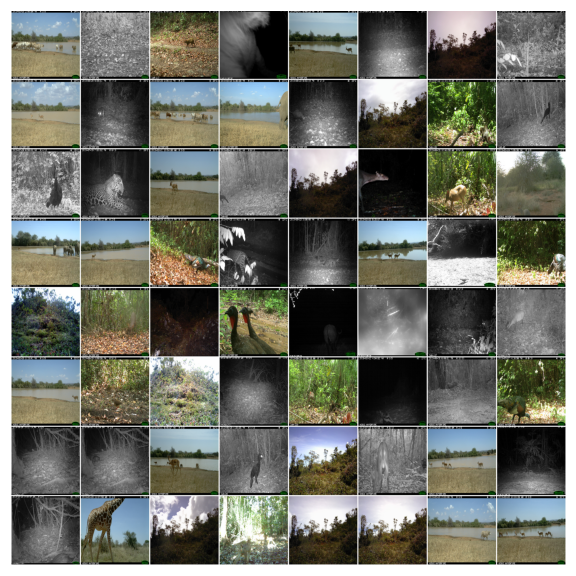}
    \caption*{\iwildcam ({\bf Shift 1: Spurious correlation.})} 
    \end{subfigure}
    
    \begin{subfigure}[b]{0.48\linewidth}
    \includegraphics[width=\linewidth]{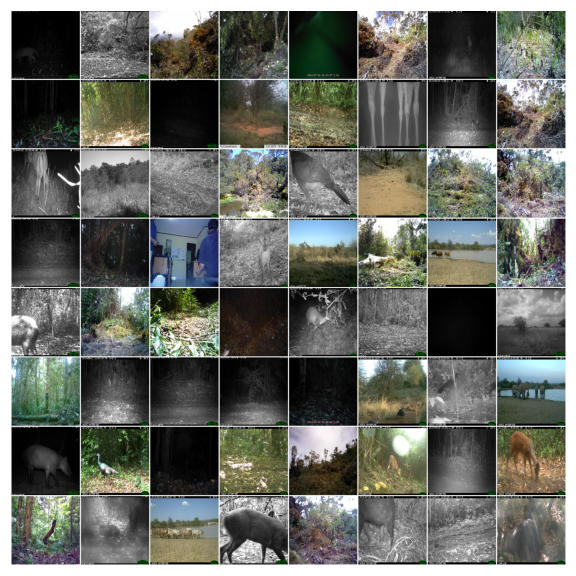}
    \caption*{\iwildcam ({\bf Shift 2: Low-data drift.})} 
    \end{subfigure}
    \begin{subfigure}[b]{0.48\linewidth}
    \includegraphics[width=\linewidth]{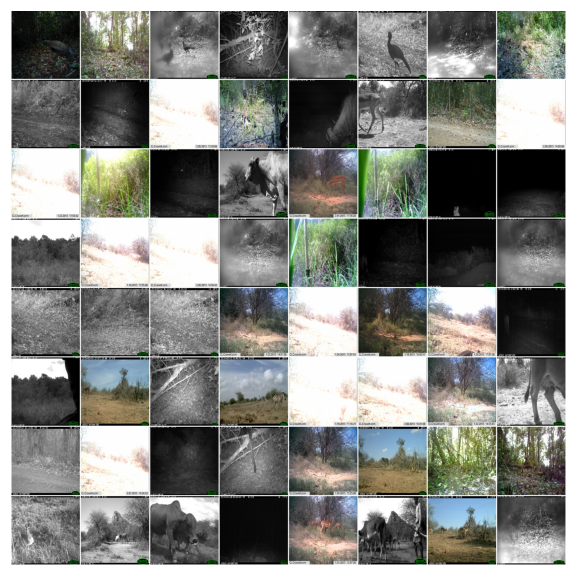}
    \caption*{\iwildcam ({\bf Shift 3: Unseen data shift.})} 
    \end{subfigure}
    \caption{Sample distributions on \iwildcam.}
    \label{fig:dataset:iwildcam}
\end{figure}

We include samples for each distribution for each dataset in \figref{fig:dataset:dsprites}-\ref{fig:dataset:iwildcam}.

\clearpage

\section{Method}
\label{sec:app:methods}
Here we give further description of the methods we implement and how they relate to our robustness framework.
This allows us to obtain an intuition of what guarantees each method gives in this context and thereby under what circumstances they should promote generalization.

\paragraph{Backbone architecture.}
We investigate the performance of different standard vision models on the robustness task.
The model is trained on $\ptrain$ to predict the true label, making no use of the additional attribute information.
We use weighted resampling $p_\textrm{reweight}$ \citep{Vapnik92} to oversample from the parts of the distribution that have a lower probability. 
This is what we refer to as the {\em standard} setup.

\paragraph{Heuristic data augmentation.}
In this case, we use standard heuristic augmentation methods in order to augment the training samples in $\ptrain$.
Instead of attempting to learn the conditional generative model, in this approach, we `fake' the generative model by augmenting the images using a set of heuristic functions to create $p_\textrm{aug}, \alpha=1$.
However, we have to heuristically choose these functions, so the generative model they approximate may {\em not} correspond to the true underlying generative model $p(\vx| \attribute{1:\NumOfAttributes})$.
In practice, these methods make no use of the additional attribute information and are trained to predict the label, as in the standard setup.

\paragraph{Learned data augmentation.}

Again, we approximate the underlying generative model $p(\vx | \attribute{1:\NumOfAttributes})$ by a set of augmentations. However, instead of heuristically choosing these augmentation functions, we learn them from data.
We learn a function that, given an image $\vx$ and attribute $y^a=v_i$, transforms $\vx$ to have another attribute value $y^a=v_j$, while keeping all other attributes fixed.
We can then use this function to generate new samples $p_\text{aug},\alpha=1$ with any distribution  over $y^a$.
In particular, we generate samples under the uniform distribution.
In this case, the additional attribute information is used to generate new samples from a given image.
However, the performance of this approach is highly dependent on the quality of the learned generative model.
We follow \cite{Goel20}, who use  \cyclegan \citep{Zhu17} to learn how to transform an image with one attribute value to have that of another. (We do not use their additional SGDRO objective as we want to study the impact of the data augmentation process alone.)
In our case, there are more than two attribute values, so we use \stargan \citep{Choi18} to learn a single model that, conditioned on an input image and desired attribute, transforms that image to have the new attribute.

\paragraph{Domain generalization.}
These works were devised to improve domain generalization.
They can also be seen as a form of representation learning. 
The aim is to recover $p(z|\vx)$ such that the representation $z$ is independent of the domain (in our framework, the attribute $y^a$): $p(y^a,z)=p(y^a)p(z)$.
If this is achieved, then the task specific classifier $p(y^l|z)$ will be independent of $y^a$ by definition.
However, these approaches rely on the ability of the underlying method to learn invariance.

\paragraph{Adaptive approaches.}
These works modify the reweighting distribution in $p_\text{reweight}$ using multi stage training.
The models are trained first as for the standard setup, giving a classifier $f_o$.
JTT \citep{Liu21} then uses $f_o$ to approximate the difficulty of the sample. 
The more difficult samples are weighted higher in the second stage according to a factor $\lambda$.
This is equivalent to sampling the more difficult sample $\lambda$ times in a batch (thereby learning a more complex function $W$).
BN-Adapt \citep{Schneider20} learns $|\attributeSet{a}|$ models in the second stage by modifying the batch normalization parameters.
For the $i$-th model, $W(y^a = v_i) = 1, W(y^a = v_j) = 0$ for $i \neq j$.
However, neither of these methods give strong guarantees on the properties of the final model.

\paragraph{Representation learning.}
Finally, in representation learning, the aim is to learn an initial representation that has preferable properties to standard ERM training; the motivation for this approach is discussed in \appref{sec:robustnessframework}.
To learn the prior, we can pretrain $f$ on large amounts of auxiliary data, such as on ImageNet \citep{Russakovsky15}.
This has been demonstrated to improve model robustness and uncertainty between datasets \citep{Hendrycks19}, but here we investigate its utility under different distribution shifts.
Another approach is to attempt to learn a disentangled representation with a VAE, as in $\beta$-VAE \citep{Higgins16}, where $z$ would then describe the underlying factors of variation for the generative model.
However, the robustness of these methods is dependent on the quality of the learned representation for the specific robustness task.

\section{Implementation}
\label{sec:app:implementation}
We first describe the architectures and precise implementation of each approach in \appref{sec:app:basearch}-\ref{sec:app:replearning} and give training details in \appref{sec:app:training}.
We then give the sweeps over the hyperparameters considered in \appref{sec:app:sweeps}.
We do not claim that these are the best possible results obtainable with each method (which would require much larger sweeps, quickly becoming computationally infeasable to compare all methods), but they are representative of performance of each approach.

\subsection{Base architectures}
\label{sec:app:basearch}

We train three types of models. These all have different capacities, which we report in \tabref{tab:app:modelcapacity}.

\paragraph{ResNets.} We use the standard ResNet18, ResNet50, and ResNet101 setups \citep{He16}.

\paragraph{MLP.} For the MLP, we use a 4 layer MLP with 256 hidden units.

\paragraph{ViT.} For the ViT \citep{Dosovitskiy20}, we set the parameters as follows.
For the smaller 64x64 images (\dsprites, \mpithreed, \shapes), we use a patch size of 4 with a hidden size of 256.
For the transformer, we use 512 for the width of the MLP, 8 heads and 8 layers.
We use a dropout rate of 0.1.
For the medium 96x96 images (\camelyon, \smallnorb), we use a patch size of 12 and for the 256x256 images (\iwildcam), a patch size of 16.

\begin{table}[t]
    \centering
    \begin{tabular}{c|c}
        {\bf Model} & {\bf \# Parameters (M)} \\ \toprule
        ResNet18 & 11.18 \\
        ResNet50 & 23.51 \\
        ResNet101 & 42.51 \\
        MLP & 3.34 \\
        ViT & 85.66 \\ \bottomrule
    \end{tabular}
    \caption{The number of parameters for each model. While ViT has the largest capacity, it was the most brittle to train and only achieved best performance on \iwildcam. On the other datasets, the ResNets performed best.}
    \label{tab:app:modelcapacity}
\end{table}

\subsection{Heuristic Augmentation}

\paragraph{ImageNet Augmentation.}
ImageNet augmentation is composed of random crops and color jitter.
We use the standard ratios as used in ImageNet training \citep{He16}. We only apply the augmentation to the first three channels of a dataset (replicating across three channels if the data is grayscale).

\paragraph{AugMix \citep{Hendrycks20b}.}
AugMix composes multiple $k$ sequences of augmentations, randomly sampled from thirteen base augmentations.
We use their default: $k=3$.

\paragraph{RandAugment \citep{Cubuk20}.}
RandAugment randomly samples $N$ augmentations from a set sixteen base augmentations with a severity $M$.
For the augmentation parameters (cutout and translate) based on image size, we interpolate between the values used for ImageNet on $224$ images and Cifar10 on $32$ images.
We set $N=3,M=5$.

\paragraph{AutoAugment \citep{Cubuk19}.}
If the image size is less than $128$ (e.g.~all datasets but \iwildcam), we use the cifar10 policy, else we use the ImageNet policy.
Again, for the augmentation parameters (cutout and translate) based on image size, we interpolate between the values used for ImageNet on $224$ images and Cifar10 on $32$ images.

\subsection{Learned Augmentation}

\paragraph{CycleGAN \citep{Goel20,Choi18,Zhu17}.}
This approach proceeds in two stages.
The first stage learns how to transform images of one attribute to have another attribute.
In this stage, these models are trained to minimize three losses: a reconstruction loss $\mathcal{L}_r$, classifier loss $\mathcal{L}_c$, and adversarial loss $\mathcal{L}_a$.
The final loss is a combination of these: $\mathcal{L} = \lambda_r \mathcal{L}_r + \lambda_c \mathcal{L}_c - \lambda_a \mathcal{L}_a$.
We set $\lambda_c = 1, \lambda_r = 1$ and we sweep over $\lambda_a$.
We train this model with a learning rate $\text{lr}_\text{\stargan}$ (using the same learning rate for the generator and discriminator).
We use the same ResNet style architecture as \cyclegan, except we append to the image a one hot encoding designating the original attribute, as described in \stargan \citep{Choi18}.
We use five ResNet blocks for images of size $64$ (\dsprites, \mpithreed, \shapes), six ResNet blocks for images of size $96$ (\camelyon, \smallnorb) and nine ResNet blocks for images of size $256$ (\iwildcam).

The second stage uses the pretrained \stargan model in order to obtain $p_\textrm{aug}$.
For each image, we uniformly at random select a value for $y^a$. We then transform the image to have the new attribute value (while keeping the label fixed).
We then use this image as an augmented sample.
We set $\alpha=0$, using no real samples.

\subsection{Domain Generalization}
\paragraph{IRM \citep{Arjovsky18}.}
IRM enforces that the optimal classifier for all domains is the same.
This is done using two terms in the risk: a term to minimize the overall risk and a term to enforce the constraint, with a tradeoff $\lambda$.

\paragraph{DeepCORAL \citep{Sun16}.}
DeepCORAL enforces that the mean and covariance of the learned features in different domains is the same.
This is achieved by two terms in the loss: a term to minimize the overall risk and a term to penalize differences in the mean and covariance across each pair of domains.
These are weighted according to a tradeoff $\lambda$.

\paragraph{Domain MixUp \citep{Gulrajani20}.}
Domain MixUp enforces smoothness in the label prediction by interpolating between images from different domains and enforcing the prediction similarly interpolates between the labels. We set the interpolation parameter $\lambda=0.2$ (as in MixUp \cite{Zhang18}).

\paragraph{DANN \citep{Ganin16}.}
DANN trains an adversary $g$ that attempts to predict the attribute from the learned representation and the model seeks to fool this adversary while also minimizing downstream accuracy: $\mathcal{L} = \mathcal{L}_c(l(z)) - \lambda_a \mathcal{L}_a(g(z))$.
For the adversary, we use a two layer MLP with ReLUs \citep{Nair10} and hidden sizes of size $64$.

\paragraph{SagNet \citep{Nam21}.}
SagNet aims to learn a representation $z$ that is invariant to style by using an adversary.
The classification loss $\mathcal{L}_c$ aims to use the content predictor $c$ to classify $z$. The adversary $\mathcal{L}_a$ aims to use the style predictor $s$ to predict the class. This gives a final loss: $\mathcal{L} = \mathcal{L}_c(c(z)) - \lambda_a \mathcal{L}_a(s(z))$.
We use the feature extractor of a ResNet18 \citep{He16} (before the average pool) as the representation $z$.
For the content $c$ and style predictor $s$, we use MLPs with three hidden dimensions of size 64 and ReLUs.
We set $\lambda_a = 1$.

\subsection{Adaptive Approaches}
\paragraph{JTT \citep{Liu21}. }
JTT uses a pretrained classifier to find the most challenging samples.
It then reweights these samples by $\lambda$.
We set $\lambda=20$ and train the first model for half the training time.

\paragraph{BN-Adapt \citep{Schneider20}. }
BN-Adapt adapts the parameters of the original model according to the attribute label using a hyperparameter $r$ ($N / n$ in their paper). Intuitively, this controls the the strength of the original model $r$ and the new model $1$.
We set $r = 10$.

\subsection{Representation Learning}
\label{sec:app:replearning}
\paragraph{$\beta$-VAE \citep{Higgins16}.}
$\beta$-VAE poses the original VAE objective using constrained optimisation to obtain a new objective which has a tradeoff $\beta$ between the log-likelihood objective and the KL objective.
We resize images to 64x64 and use a convolutional architecture as in \cite{Higgins16}.
The encoder consists of the following layers (with ReLUs) to obtain the final representation $z$ of size $L$: Conv 16x5x5 (stride 1, spatial feature size: 60), Conv 32x4x4 (stride 2, spatial feature size 29), Conv 64x3x3 (stride 1, spatial feature size 27), Conv 128x3x3 (stride 2, spatial feature size 13), Conv 256x4x4 (stride 1, spatial feature size 10), Conv 512x4x4 (stride 2, spatial feature size 4), Conv 512x4x4 (stride 1, spatial feature size 1), Linear $L$.
The decoder has the reverse setup to the encoder with convolutional transpose layers.
The learned representation is kept fixed when training the downstream classifier.

There are two terms to trade-off: the strength of $\beta$ and the size of the latent representation $L$.
However, a larger latent will need a stronger $\beta$.
We use the $\beta_{\text{norm}} = \frac{\beta L}{I}$ which normalises these values by the data size $I$.
We then sweep over $L$ and $\beta_{\text{norm}}$.

Finally, to train the downstream classifier, we fix the encoder of the $\beta$-VAE. 
We then train an MLP (the same as used in \appref{sec:app:basearch}) from the representation of size $L$ to the labels.

\paragraph{Pretrained on ImageNet.}
We resize images to 224x224 and pass them to a model pretrained on ImageNet. We finetune the full network.

\subsection{Training.}
\label{sec:app:training}
We train the ResNets and MLP with the Adam optimizer for a maximum of 100K steps on the synthetic datasets and 200K steps on \camelyon and \iwildcam with a batch size of 128.
The ViT is trained with a batch size of 1024 (it did not converge for smaller batch sizes).
All models are trained with early stopping using the validation accuracy.

\subsection{Sweeps.}
\label{sec:app:sweeps}
The sweeps are given in \tabref{tab:app:sweeps}.

\begin{table}[ht]
    \scriptsize
    \centering
    \begin{tabular}{cp{5cm}}
         {\bf Method} & {\bf Sweeps}\\ \toprule
         
         ResNets & learning rate: [1e-2, 1e-3, 1e-4] \\  \midrule
         MLP & learning rate: [1e-2, 1e-3, 1e-4] \\  \midrule
         ViT & learning rate: [1e-2, 1e-3, 1e-4] \\  \midrule  \midrule
         ImageNet Augm & learning rate: [1e-2, 1e-3, 1e-4] \\  \midrule
         AugMix & learning rate: [1e-2, 1e-3, 1e-4] \\  \midrule
         RandAugment & learning rate: [1e-2, 1e-3, 1e-4] \\  \midrule
         AutoAugment & learning rate: [1e-2, 1e-3, 1e-4] \\  \midrule  \midrule 
         \cyclegan & learning rate: [1e-4] \newline $\lambda_a$: [0.01, 0.1, 1, 10] \newline $\text{lr}_\text{\stargan}$: [1e-3, 1e-4, 1e-5] \\ \midrule \midrule
         IRM & learning rate: [1e-2, 1e-3, 1e-4] \newline $\lambda$: [0.01, 0.1, 1, 10] \\ \midrule
         MixUp & learning rate: [1e-2, 1e-3, 1e-4] \\  \midrule
         CORAL & learning rate: [1e-2, 1e-3, 1e-4] \newline $\lambda$: [0.01, 0.1, 1, 10] \\ \midrule
         SagNet & learning rate: [1e-2, 1e-3, 1e-4] \\ \midrule
         DANN & learning rate: [1e-2, 1e-3, 1e-4] \newline $\lambda_a$: [0.01, 0.1, 1, 10] \\ \midrule \midrule
         JTT & learning rate: [1e-2, 1e-3, 1e-4] \\  \midrule
         BN-Adapt & learning rate: [1e-2, 1e-3, 1e-4] \\  \midrule \midrule
         $\beta$-VAE & learning rate: [1e-4] \newline $\log(\beta_\text{norm}): $ linspace(1e-4,10,4) \newline $L:$ linspace(10,210,4) \\  \midrule
         Pretrained on ImageNet & learning rate: [1e-2, 1e-3, 1e-4] \\  \midrule
         
    \end{tabular}
    \caption{The sweeps over each method for each of the five seeds. The maximum total sweep size (due to capacity) is eight models per seed. For each seed, the best model (based on the ID or OOD validation set) is selected from the hyperparameter sweep and used at evaluation to compute the test time performance.}
    \label{tab:app:sweeps}
\end{table}

\end{document}